\documentclass[10pt,twocolumn,letterpaper]{article}

\usepackage{iccv}
\usepackage{times}
\usepackage{epsfig}
\usepackage{graphicx}
\usepackage{amsmath}
\usepackage{amssymb}
\usepackage{multirow}
\usepackage{multicol}
\usepackage{caption}
\usepackage{booktabs}
\usepackage{subcaption}
\usepackage{tabularx}
\usepackage{amsfonts}
\usepackage{cite}
\usepackage{threeparttable}
\usepackage{slashbox}

\usepackage[breaklinks=true,bookmarks=false]{hyperref}

\iccvfinalcopy 

\def\iccvPaperID{****} 

\ificcvfinal\fi

\begin{document}

\title{DeepVAT: A Self-Supervised Technique for Cluster Assessment in Image Datasets}

\author{Alokendu Mazumder\textsuperscript{1}, Tirthajit Baruah\textsuperscript{2}, Akash Kumar Singh\textsuperscript{2},\\ Pagadla Krishna Murthy\textsuperscript{1}, Vishwajeet Pattanaik\textsuperscript{1}, and  Punit Rathore\textsuperscript{1}\\
\textsuperscript{1}Indian Institute of Science, Bengaluru, India\\
\textsuperscript{2}Indian Institute of Science Education and Research, Bhopal, India\\
{\tt\small \{alokendum, pagadlam, vishwajeetp, prathore\}@iisc.ac.in}
\\
{\tt\small \{tirthajitoff, akashadarsh21\}@gmail.com}
}

\maketitle
\ificcvfinal\thispagestyle{empty}\fi

\begin{abstract}
Estimating the number of clusters and cluster structures in unlabeled, complex, and high-dimensional datasets (like images) is challenging for traditional clustering algorithms. In recent years, a matrix reordering-based algorithm called \textit{Visual Assessment of Tendency} (VAT), and its variants have attracted many researchers from various domains to estimate the number of clusters and inherent cluster structure present in the data. However, these algorithms face significant challenges when dealing with image data as they fail to effectively capture the crucial features inherent in images.
To overcome these limitations, we propose a deep-learning-based framework that enables the assessment of cluster structure in complex image datasets.
Our approach utilizes a self-supervised deep neural network to generate representative embeddings for the data. These embeddings are then reduced to 2-dimension using t-distributed Stochastic Neighbour Embedding (t-SNE) and inputted into VAT based algorithms to estimate the underlying cluster structure. Importantly, our framework does not rely on any prior knowledge of the number of clusters. Our proposed approach demonstrates superior performance compared to state-of-the-art VAT family algorithms and two other deep clustering algorithms on four benchmark image datasets, namely MNIST, FMNIST, CIFAR-10, and INTEL.

\end{abstract}

\section{Introduction}
Data clustering is a widely used unsupervised learning technique that involves dividing a collection of unlabeled objects into $k$ groups of similar objects. Various clustering algorithms are available in the literature, such as hierarchical clustering, centroid-based approaches, density-based algorithms, and distribution-based clustering. Most clustering algorithms require $k$, the number of clusters to seek, as an input, which is the clustering tendency assessment problem. One common method to determine the number of clusters and their underlying structure is to visualize the data points using a 2D or 3D plot. However, this approach is only feasible for two- or three-dimensional datasets. For high-dimensional datasets such as images, time-series, visualizing and interpreting cluster structures using 2D or 3D visualization is not practical. Although various dimensionality reduction techniques, such as principal component analysis (PCA) and linear discriminant analysis (LDA), exist in the literature, these techniques often result in a low-dimensional representation of complex, high-dimensional datasets that may not fully reflect the inherent cluster structure due to information loss.

There are various formal (based on statistics) and informal (other approaches) techniques~\cite{jain1988algorithms,everitt1978graphical}  available in the literature for cluster structure assessment, but they are not completely effective. In contrast, visual approaches~\cite{cleveland1993visualizing}  have been in use for many years and serve as the foundation for many visual data analysis methods. The \textit{Visual Assessment of Clustering Tendency} (VAT)~\cite{bezdek2002vat},  a matrix reordering-based visual-analytical method, is one of such algorithm which provides a visual way to assess the clustering tendency of various datasets. There are several variants of VAT available for different types of data, which are collectively known as the VAT family of algorithms. The VAT family of algorithms has become an acceptable and widely used tool in several domains like biomedical applications, speech processing, image segmentation, transportation applications, and \textit{etc} for exploratory data analysis. 

VAT algorithm employs a variant of Prim's minimum spanning tree algorithm~\cite{prim1957shortest} to perform matrix reordering of the pairwise dissimilarity matrix to generate a reordered dissimilarity matrix. The reordered dissimilarity matrix can be viewed as a monochrome image called a \textit{Reordered Dissimilarity Image} (RDI) or cluster heat map. The RDI displays a possible cluster structure of the data set by showing dark blocks (data points of low dissimilarity values) along the diagonal.  One method to obtain an accurate estimate of the number of clusters ($k$) from the RDI in the data is to count the number of dark blocks along the diagonal of the RDI. That means VAT not only can be used for cluster tendency assessment but also can be used for subsequent clustering of the input datasets, without needing the number of clusters.

This method is particularly effective for datasets with well-separated, compact clusters since the dark blocks along the diagonal are easily identifiable. However, for complex datasets (e.g., images, time series) having overlapping cluster structures (which is the case for most real-life datasets), existing VAT approaches perform poorly as the RDI quality degrades and the contrast between dark blocks along the diagonal and the rest of the image decrease. This makes it difficult to count the dark blocks along the diagonal.

\begin{figure}[t]
\captionsetup[subfigure]{justification=centering}
\centering
\subfloat[]{\includegraphics[width=0.1\textwidth]{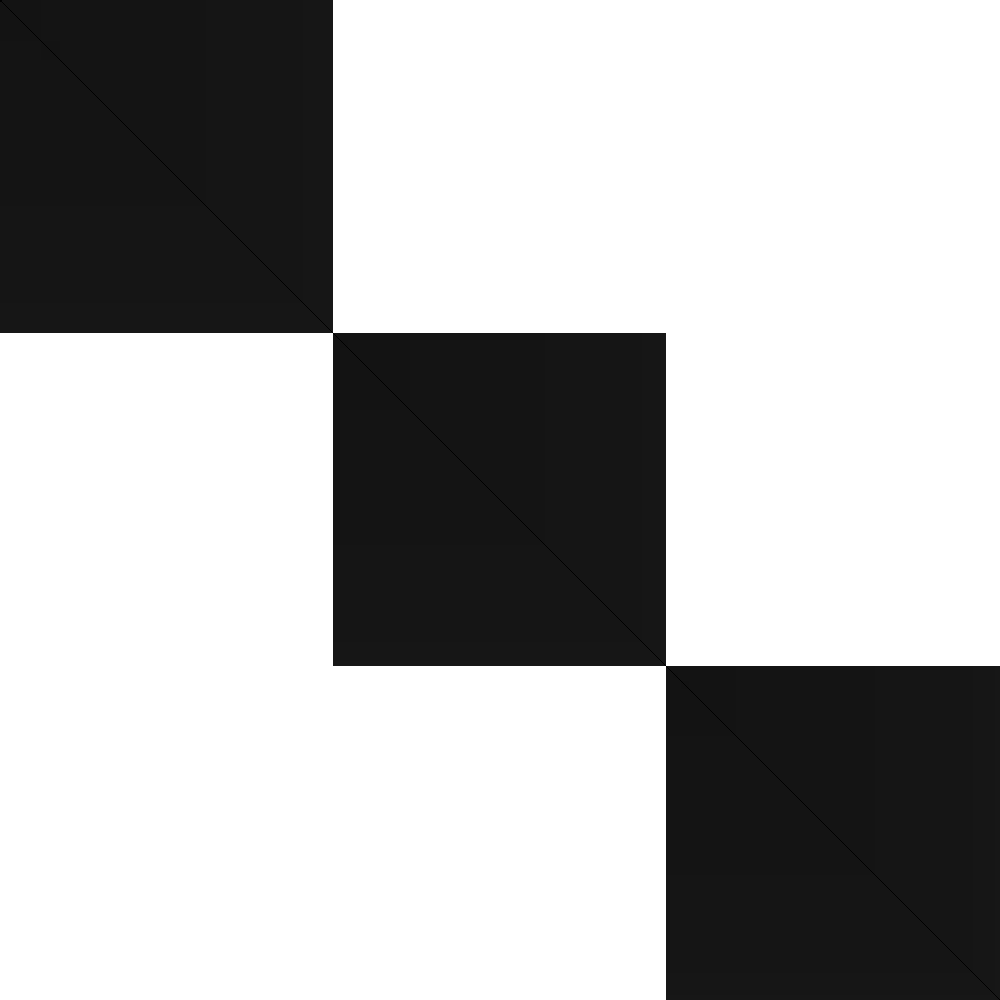}} \hspace{10ex}
\subfloat[]{\includegraphics[width=0.1\textwidth]{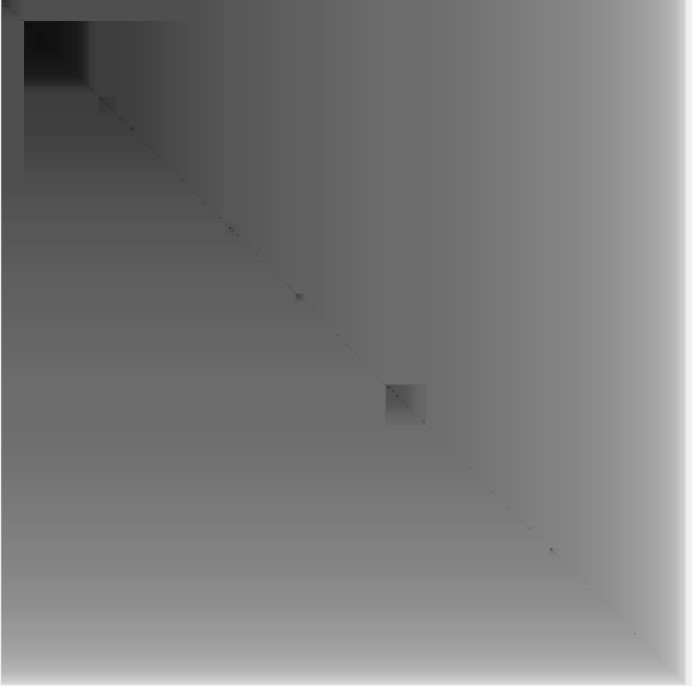}}
\caption{(a) iVAT image of 100-dimensional Compact and Separated (CS) Gaussian mixture data (3 Gaussians); (b) iVAT image of flattened MNIST data (784 dimensional)}%
\label{Fig:ivat_mnist}
\end{figure}

There have been some efforts~\cite{havens2011efficient, rathore2018rapid, PPR:PPR472596} to improve the quality of VAT generated RDI to accurately estimate the number of clusters for various complex geometry datasets. The VAT family algorithms, commonly used for analyzing cluster structures, exhibit poor performance when applied to image datasets, especially those with overlapping clusters. In the typical workflow, images are flattened before employing the VAT algorithms, resulting in the loss of their crucial spatial features. Consequently, the pixel-wise Euclidean distance becomes less effective in accurately capturing similarities or dissimilarities between images due to the feature loss incurred during flattening and the curse of dimensionality. Figure ~\ref{Fig:ivat_mnist}  shows an \textit{improved Visual Assessment of Tendency} (iVAT) \cite{havens2011efficient} RDI for a synthetic, high-dimensional dataset (number of samples = 1000, dimensions= $100$) having three well-separated Gaussian mixtures (so $k$=3) in View (a), and RDI for a sample of popular MNIST dataset (number of samples = 1000 dimensions= $784$, $k=10$ classes) in View (b). It is evident from the figure that iVAT performs well when the data has inherently well separated clusters as we can clearly see three dark blocks along the diagonal in its RDI representing three clusters. However, when it comes to image datasets like MNIST, it struggles to provide meaningful results, as the resulting RDI does not exhibit clear dark blocks along the diagonal. This limitation highlights the need for a VAT variant that can effectively preserve the essential features of images, enabling more accurate assessments of cluster structures in image datasets.

As unsupervised deep learning methods (like autoencoders~\cite{vincent2010stacked} and contrastive learning~\cite{chen2020simple}) excel at extracting robust features from complex images, it makes them well-suited for developing a dedicated VAT algorithm for image datasets.

To address the above concerns, we propose a novel visual-analytical framework called DeepVAT. DeepVAT utilizes deep learning techniques to extract meaningful deep features from images, enabling more effective assessment of cluster structures. Unlike traditional VAT approaches, DeepVAT can uncover hidden cluster structures within image data, even in situations where ground truth labels or information about the number of classes are unavailable. 
Our major contributions are as follows:
\begin{enumerate}
    \item We proposed a deep, self-supervised learning framework, DeepVAT, that can provide visual evidence of the number of clusters present in complex image datasets. 
     
    \item In our method, we did not incorporate any prior knowledge about the ground truth number of clusters of data.
    \item We performed experiments on four real-world, publicly available, large image datasets to show the superiority of DeepVAT over other state-of-the-art VAT family algorithms (proposed for high-dimensional data) in terms of quality of RDI, clustering accuracy, and normalized mutual information (NMI) score.
\end{enumerate}

To the best of our knowledge, our work represents the first investigation in the literature exploring the utilization of deep features from images in the context of VAT methods. This contribution highlights the importance of incorporating deep learning techniques in the development of VAT models for accurate and insightful analysis of image datasets.

Here is an outline of the rest of this article. Section~\ref{sec:prelim} presents the preliminaries for the VAT/iVAT algorithm and reviews related work. The proposed algorithm, DeepVAT, is discussed in Section~\ref{sec:DeepVAT}. Section~\ref{sec:exp} discusses the experiments and results, followed by conclusions in Section~\ref{sec:conc}.





\begin{figure*}
    \centering
    \includegraphics[width=0.9\textwidth ]{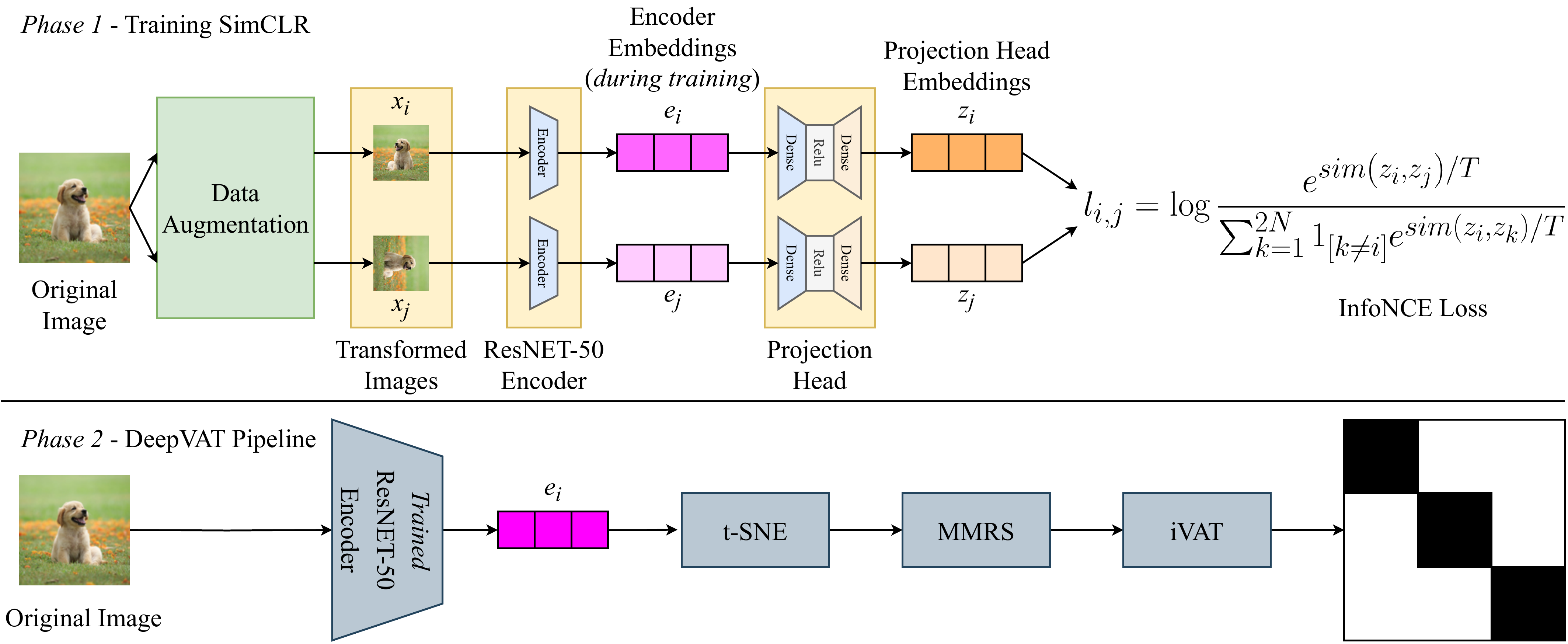}
    \caption{The proposed architecture of DeepVAT}
    \label{fig:method}
\end{figure*}

\section{Preliminaries and Related work}
\label{sec:prelim}
\subsection{VAT and iVAT}
\label{sec: vat_ivat}
Consider we have a set of $N$ objects, denoted as $O={o_{1},o_{2},\ldots,o_{N}}$, where each object in $O$ is described by a $p$-dimensional feature vector ($\in \mathbb{R}^{p}$). 
Alternatively, the data can be represented as a dissimilarity matrix, denoted as $D_N = [d_{ij}]$, where $d_{ij}$ indicates the dissimilarity between object $o_{i}$ and object $o_{j}$ computed using a suitable distance measure. The VAT algorithm considers the dissimilarity matrix, $D_N$ as input and reorders (by shuffling the rows and columns) using a modified Prim's algorithm. The image $I(D_N^{})$ of the reordered distance matrix $D_N^{}$ displays each pixel's intensity to indicate the dissimilarity between the corresponding row and column objects. When dark blocks appear along the diagonal, they might represent distinct clusters, ideally $k$ (the original number of clusters in data) clusters. As single-linkage clusters are always diagonally aligned in VAT ordered images~\cite{havens2013scalable}, $k_{p}$ aligned clusters can be obtained by cutting the largest $(k_{p}-1)$ edges (given by the MST cut magnitude order $d$) from the MST. Here, $k_{p}$ is the estimated number of clusters from VAT/iVAT RDI.

The improved-VAT (iVAT)~\cite{havens2011efficient} enhances the quality of VAT~\cite{bezdek2002vat} RDI by using path-based distance transformation. The iVAT transformed matrix $D_N^{'}= [d_{ij}^{'}]$ is generated using a path-based minimax distance~\cite{prim1957shortest}:
\begin{equation}
    d^{'}_{ij} = \min\limits_{p \in P_{ij}} \max\limits_{1 < h < |p|} \textbf{D}_{p[h]p[h+1]}
\end{equation}
where $p \in P_{ij}$ is an acyclic path in the set of all acyclic paths between objects $(o_{i})$ and  $(o_{j})$ (vertices $i$ and $j$) in $O$.


\subsection{VAT Variants for Large Volumes of High-Dimensional Data}
Although the VAT tool, discussed above, finds its usefulness in many applications, it can be computationally expensive as the size of the data set grows due to its $\mathcal{O}(N^2)$ complexity. To understand the clustering structure for large volume datasets, a scalable version of VAT called \emph{scalable VAT} (sVAT) was developed by Hathaway \textit{et al.}~\cite{hathaway2006scalable}, which utilizes a smart sampling based approach. To begin, sVAT extracts a smart sample of size $n$ (where $n<<N$) from the large data set $X$ using \emph{Maximin Random Sampling} (MMRS)~\cite{johnson1990minimax}. The extracted sample is then used to compute the distance matrix $D_n$, which is input into VAT.

To handle large volumes of high-dimensional datasets, Rathore \textit{et. al} in~\cite{rathore2018rapid} proposed FensiVAT, an ensemble-based, hybrid clustering framework that combines fast data-space reduction using random projection with an intelligent sampling strategy to assess the clustering tendency of high-dimensional data.  Recently, Zhang \textit{et al.}~\cite{PPR:PPR472596} proposed another method that leverages a kernel-based dissimilarity matrix to refine the RDI further, called kernel-based iVAT (KernelVAT). They use a Gaussian kernel and Isolation kernel (data-dependent) to transform the RDI. 

The SpecVAT~\cite{wang2008specvat} algorithm is another approach that improves the quality of the RDI produced by VAT. It utilizes spectral graph theory to transform the raw distance matrix into a graph embedding space using graph Laplacian. It then creates an alternative feature representation of the data by selecting the $r$ most significant eigenvectors that correspond to the highest eigenvalues. VAT is then applied to this transformed representation, resulting in a much-improved RDI.

To our knowledge, none of the existing VAT family of algorithms, including those reviewed in this section, have been investigated thoroughly on image datasets. Moreover, they have been shown to perform poorly on image datasets in their numerical experiments. Below, we discuss our proposed framework, DeepVAT.





\section{Proposed Framework: DeepVAT}
\label{sec:DeepVAT}
In this paper, we propose a deep learning-based framework, DeepVAT, to advance the VAT family of algorithms for cluster structure assessment in complex image datasets. Figure ~\ref{fig:method} presents each step of our proposed framework. Below, we briefly explain each step of DeepVAT keyed to the blocks shown in figure ~\ref{fig:method}.

\subsection{Generating Image Embeddings}
The first step in our framework is representation learning by employing deep learning architectures. The objective of this step is to attain a \emph{cluster-friendly} representation of images, which involves bringing similar data points closer to each other and pushing dissimilar points further away. While deep learning architectures like autoencoders can be explored for this purpose, they may lack the inherent capability to produce a truly \emph{cluster-friendly} representation. 

Recently, a wide range of self-supervised approaches such as contrastive learning based models has been proposed that can provide \emph{cluster-friendly} representations for images using deep neural networks, without the need for ground truth information. These models include \textit{Simple Contrastive Learning of Representations} (SimCLR)~\cite{chen2020simple}, Barlow Twins~\cite{zbontar2021barlow}, \textit{Decoupled Contrastive Learning} (DCL)~\cite{avidan2022computer}, SimSiam~\cite{chen2021exploring}, \textit{Bootstrap Your Own Latent} (BYOL)~\cite{grill2020bootstrap}, and many others.

We observed that incorporating SimCLR as the feature extractor in the DeepVAT pipeline led to significantly superior iVAT images compared to when an autoencoder was used as the feature extractor. The significant difference observed can be attributed to the inherent capabilities of SimCLR compared to basic autoencoders. SimCLR has the ability to effectively group similar points together and push dissimilar points apart, thanks to the InfoNCE loss it minimizes~\cite{oord2018representation}. In contrast, basic autoencoders lack this inherent capability. Additionally, a recent study~\cite{parulekar2023infonce} suggests that the InfoNCE loss aids in learning cluster-preserving representations of images, further highlighting the suitability of SimCLR for DeepVAT.
Hence, we chose SimCLR as our primary model for creating embeddings in our proposed framework. 




In SimCLR, the first stage includes performing auxiliary tasks or a given batch of images, such as corrupting the data, adding noise, and creating augmented views of the same data. 
These transformations generate fresh views of the same images, effectively enlarging the training set. Gray-scale images cannot undergo certain transformations such as color jitters. Instead, an affine stretch is utilized along with rotation, resizing, and blurring. Through these tasks, the models can acquire a rich and beneficial representation of the data.

SimCLR consists of an encoder network and a non-linear projection head. The augmented images are fed into the encoder to extract high-level features. The encoder consists of several convolutional and fully connected layers and is trained using a contrastive loss function. The SimCLR framework utilizes an InfoNCE loss function~\cite{oord2018representation} to measure the similarity between different views of an image. The model aims to maximize the similarity between the two views of the same image and minimize the similarity between views of different images. By doing so, SimCLR learns to extract valuable features robust to variations in the input data, which is helpful for generalization in real-world scenarios. The encoder projects the images into (say) $d$-dimensional space. 

Then, the projection head, a small neural network, further maps the encoded features ($d$-dimensional) to a (lower) $m$-dimensional set of embeddings, and then back to a lower-manifold of $d$-dimensional space, resulting in a rank-deficient weight matrix. This projection head is trained alongside the encoder during training. After training successfully, the projection head is discarded, and the data is passed through the trained encoder to generate embeddings. 
The projection head serves as an additional non-linear transformation that helps to increase the quality of the learned features. 


\subsection{Dimensional Reduction using t-SNE}
Despite the fact that SimCLR embeddings (shown with a pink bar in figure ~\ref{fig:method}) can be used to compute the dissimilarity matrix for VAT/iVAT, the high dimensionality of the SimCLR embeddings can lead to the curse of dimensionality problem, which can affect the quality of the resulting visualization. In our experiments, as discussed in Section~\ref{sec:exp}, we observed that using SimCLR embeddings to compute the dissimilarity matrix did not result in a significant improvement in the quality of the resulting RDI for complex image datasets (CIFAR-10~\cite{cifar10} and INTEL~\cite{intelimages}).

One way to tackle this issue is to apply t-SNE on a data representation obtained from SimCLR. Compared to the original flattened image data, t-SNE works better on SimCLR embeddings because SimCLR is a deep-layer architecture that can more efficiently represent the highly varying data manifold in multiple nonlinear layers~\cite{chen2020simple,van2008visualizing}. 
The projections generated by SimCLR's projection head can identify highly varying manifolds better than a local method like t-SNE, resulting in a higher quality visualization compared to using t-SNE on the original high-dimensional data~\cite{van2008visualizing}. However, it is important to acknowledge that representing the complete structure of intrinsically high-dimensional data in just two or three dimensions is fundamentally impossible, highlighting a fundamental limitation.

\subsection{Smart Sampling: Maximin Random Sampling (MMRS)}
Computing and analyzing VAT RDI using t-SNE embeddings (shown with a pink bar in figure ~\ref{fig:method}), generated in the last step, may be infeasible for image datasets with large samples $(N)$ due to $\mathcal{O}(N^2)$ complexity of VAT. To deal with large image datasets, we exploit a smart sampling approach called \textit{Maximin and Random Sampling }(MMRS). 

Let $\textbf{X} = \{x_{i}\}_{i=1}^{N}$ represent the set of t-SNE reduced embeddings obtained from the trained encoder, where $x_{i} \in \mathbb{R}^{2}$. The MMRS technique is an intelligent way to obtain samples in large batch data sets by combining MaxiMin (MM) and Random Sampling (RS). The MM sampling process starts by identifying a set of $k^{'}$ (an overestimate of $k$) distinguished objects, which are the farthest from each other in the input data $\textbf{X}$. Then each point in the set $\textbf{X}$ is grouped with its nearest distinguished object using the nearest prototype rule (NPR) (mentioned in ~\cite{rathore2018rapid}), which divides the entire dataset into $k'$ groups  $\{G_{i}\}_{i=1}^{k^{'}}$ where $G_{i} \subseteq \textbf{X}$, $\forall i \in \{1,2,\dots,k^{'}\}$ by associating $|G_{i}|$ points to $i^{th}$ \textit{MM sample}, which represents each of the $k^{'}$ group. Finally, the sample $\mathcal{S}$ of size $n << N$ is formed by selecting random data-points from each of the $k^{'}$ groups $\{G_{i}\}_{i=1}^{k^{'}}$. The number of points $n_{j}$ extracted from group $G_{j}$ is proportional to the cardinality of $G_{j}$, i.e $n_{j} \propto |G_{j}|$. To be precise, $n_{j} = \lceil n \times |G_{j}|/N \rceil$, where $\lceil . \rceil$ is the ceiling function. This step gives us a smart sample of size $n<<N$ in lower dimensional space. Rather than feeding a large number of embeddings directly into iVAT for visualization, we feed a \textit{smart sample} of size $n$, obtained using MMRS. 

\subsection{Dissimilarity Matrix Computation for VAT/iVAT}
The reduced-dimension, smart samples are used to compute dissimilarity matrix $D_n$ which is fed to the VAT/iVAT algorithm to obtain reordered dissimilarity matrix $D_n^{'}$.
The visualization of $I(D_n^{'})$ suggests the number of clusters $k$ present in the dataset. 

\section{Experiments}
\label{sec:exp}
We performed experiments on four publicly available, real, image datasets. We evaluated the ability of DeepVAT to suggest the number of clusters in image datasets and compared its performance with other VAT family methods that are claimed to work better with high-dimensional data. We also compare DeepVAT with two well known deep-clustering based methods.  The experiments were conducted on a regular PC with the following configuration: OS: Ubuntu $22.04.2$ LTS (64 bit); processor: Intel(R) Xeon(R) Gold $5220$R CPU @ $2.20$GHz; RAM: $62$ GB; GPU: Nvidia Quadro RTX $6000$, $24$ GB.

\subsection{Datasets}
We performed our experiments on the following datasets:
\begin{enumerate}
    \item \textbf{MNIST}~\cite{lecun-mnisthandwrittendigit-2010}: It has a total of $60,000$ grayscale training images of digits with a dimension of $28*28$ ranging from 0 to 9, i.e., total $10$ classes, with each class having 6,000 images. The full training set is used in all experiments (60,000 images). 
    \item \textbf{FMNIST}~\cite{xiao2017fashion}: It has a total of 60,000 grayscale training images of fashion apparel with a dimension of $28*28$, i.e., it has a total of $10$ classes, with each class having 6,000 images. The full training set is used in all experiments (60,000 images). 
    \item \textbf{CIFAR10}~\cite{cifar10}: It has a total of $50,000$ natural RGB training images with a dimension of $32*32*3$. It has a total of $10$ classes, with each class having $5,000$ images. The full training set is used in all experiments (50,000 images). 
    \item \textbf{Intel Image Dataset}~\cite{intelimages}: It has $14,034$ natural RGB training images and 3,000 testing images with $6$ classes. We clubbed both sets and used the final count of $17,000$ images to perform various experiments. Each image has a dimension of $32*32*3$.
\end{enumerate}

\subsection{Evaluation Criteria}
We show all (best) iVAT images with an estimated number of clusters ($k_p$) for all the compared algorithms in Table~\ref{tab:resultsC}. To estimate $k_p$, we used the algorithm presented in~\cite{wang2009automatically}. As mentioned in section~\ref{sec: vat_ivat}, \emph{$k_{p}$} clusters can be obtained by cutting \emph{($k_{p}$-1)} edges in MST provided by VAT/iVAT algorithm. We used the predicted labels and ground truth information of each dataset to compute the partition accuracy (PA) for the estimated value of $k$ (from iVAT image) and normalized mutual information (NMI). The PA of a clustering algorithm is the percentage (\%) ratio of the number of samples with matching ground truth and algorithmic labels to the total number of samples in the dataset.  To ensure consistent label mapping between the predicted and true labels, the Kuhn-Munkres algorithm \cite{lovasz1986matching} is employed to find the best mapping between the predicted and ground truth labels. A higher value of PA and NMI implies a better match to the ground truth partition.
\subsection{Comparison of DeepVAT with other Models}
\label{sec: comparison}
In this section, we make a qualitative and quantitative comparison of DeepVAT with existing state-of-the-art VAT family methods that claim to work with high-dimensional and complex data (images when flattened can be seen as high-dimensional data). Specifically, we compare DeepVAT with the following methods:

\begin{enumerate}
    \item \textbf{VAT family methods}
    \begin{enumerate}
        
    \item \textbf{FensiVAT}: FensiiVAT~\cite{rathore2018rapid}  is applied on a small MMRS subset of the embeddings extracted from the trained encoder of SimCLR.
    \item \textbf{KernelVAT}: KernelVAT~\cite{PPR:PPR472596}. is applied on a small MMRS subset of the embeddings extracted from the trained encoder of SimCLR. 
    \item  \textbf{SpecVAT}: SpecVAT~\cite{wang2008specvat} is applied on a small MMRS subset of the embeddings extracted from the trained encoder of SimCLR. 
    \end{enumerate}
    \item \textbf{Deep-Clustering methods}
    \begin{enumerate}
        \item \textbf{DEC}~\cite{pmlr-v48-xieb16}: iVAT is applied to the t-SNE reduced embeddings, extracted from the trained encoder of DEC. Specifically, iVAT is applied to a smaller MMRS subset.
        \item \textbf{LSD-C}~\cite{Rebuffi_2021_ICCV}: The t-SNE reduced embeddings, extracted from the trained encoder of DEC, are utilized for applying iVAT. More specifically, iVAT is applied to a smaller MMRS subset.
        \item \textbf{Autoencoder + iVAT}: We trained a vanilla autoencoder and obtained embeddings from the trained encoder network. Subsequently, t-SNE is applied to these embeddings, and iVAT is then applied specifically to a smaller MMRS subset of the reduced embeddings.
    \end{enumerate}

\end{enumerate}

\begin{table*}
\centering
\caption{\textbf{Qualitative comparison with other methods.} When considering the visual quality of the RDI, our method stands out by providing the most accurate estimate of the number of clusters. \textbf{Note: Columns 2, 3, and 4:} FensiVAT, KernelVAT, and SpecVAT are applied to 2048-dimensional embeddings from SimCLR. \textbf{Columns 5 and 6:} iVAT is applied to the t-SNE reduced embeddings from the encoder of DEC and LSD-C, respectively. \textbf{Column 7:} iVAT is applied to the t-SNE reduced embeddings from the encoder of the autoencoder.}
\label{tab:resultsC}
\begin{tabular}{|c|c|c|c||c|c||c||c|}
\hline
\textbf{Dataset} & \textit{FensiVAT} & \textit{KernelVAT} & \textit{SpecVAT} & \textit{DEC} & \textit{LSD-C} & \textit{Autoencoder} & \textbf{Ours} \\
\hline
\textit{MNIST} &
\raisebox{-0.5\height}{\includegraphics[scale=0.2]{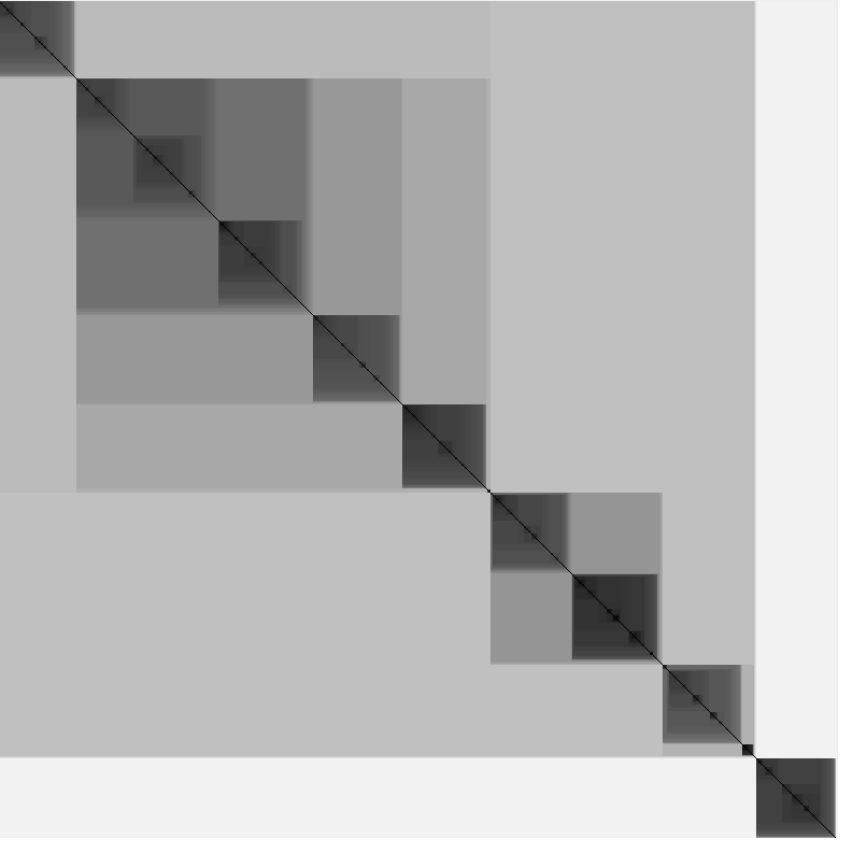}} &
\raisebox{-0.5\height}{\includegraphics[scale=0.2]{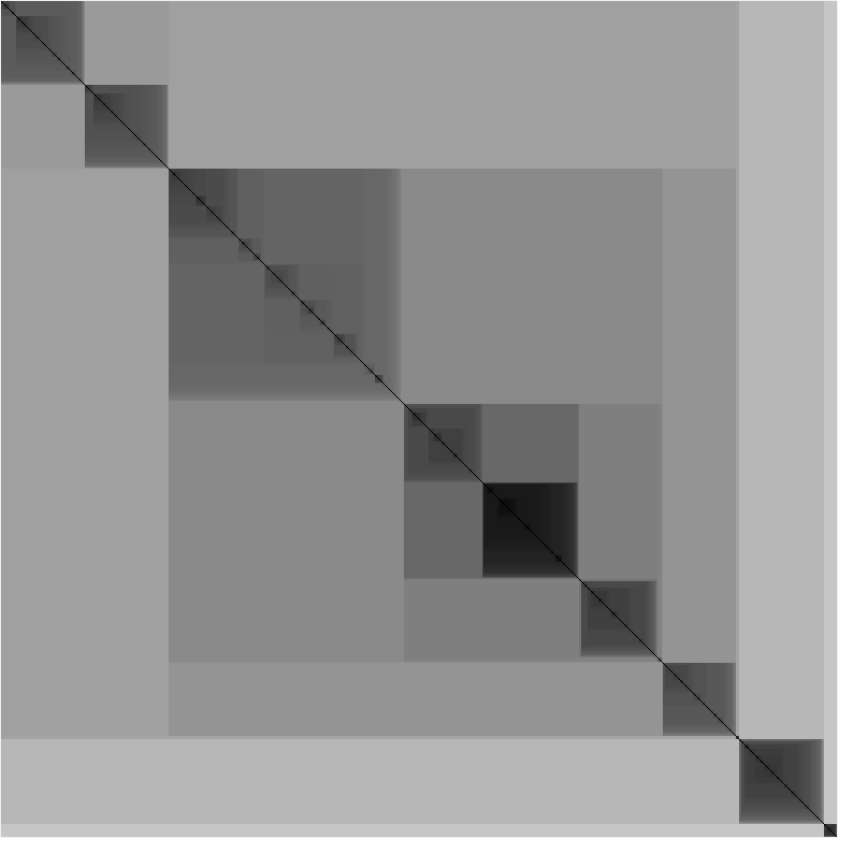}} &
\raisebox{-0.5\height}{\includegraphics[scale=0.2]{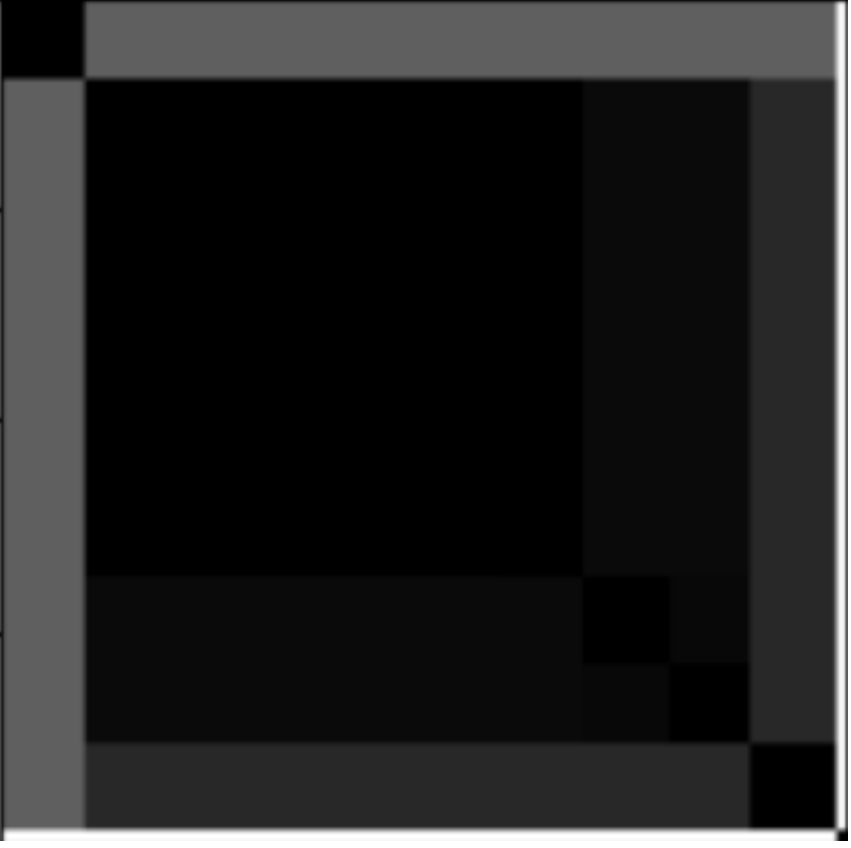}} &
\raisebox{-0.5\height}{\includegraphics[scale=0.2]{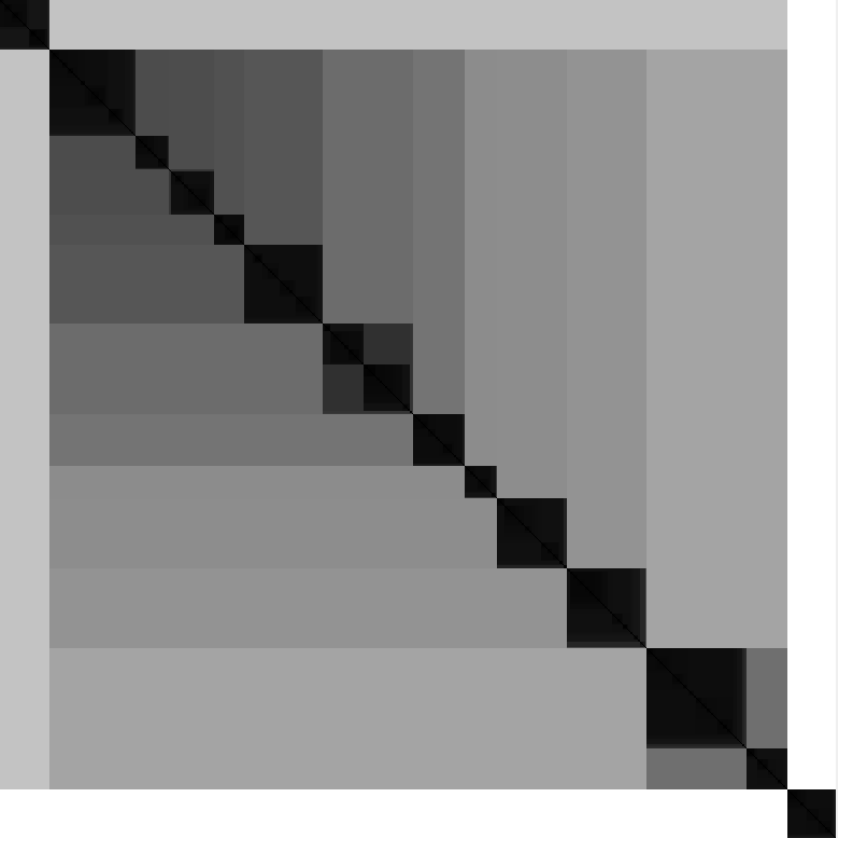}} &
\raisebox{-0.5\height}{\includegraphics[scale=0.2]{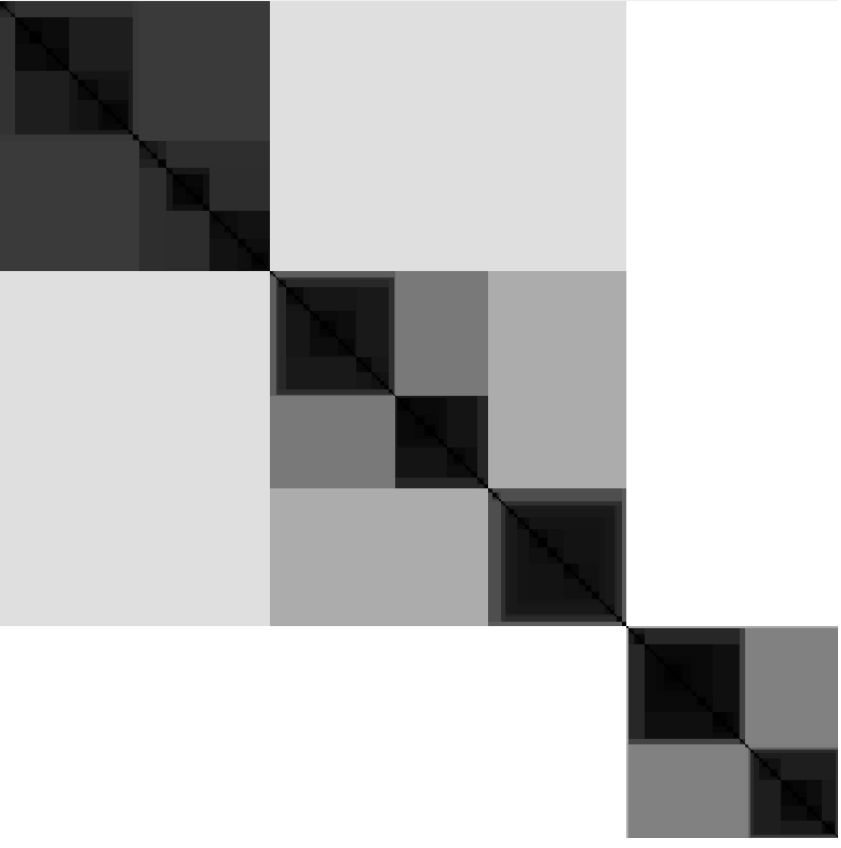}} &
\raisebox{-0.5\height}{\includegraphics[scale=0.2]{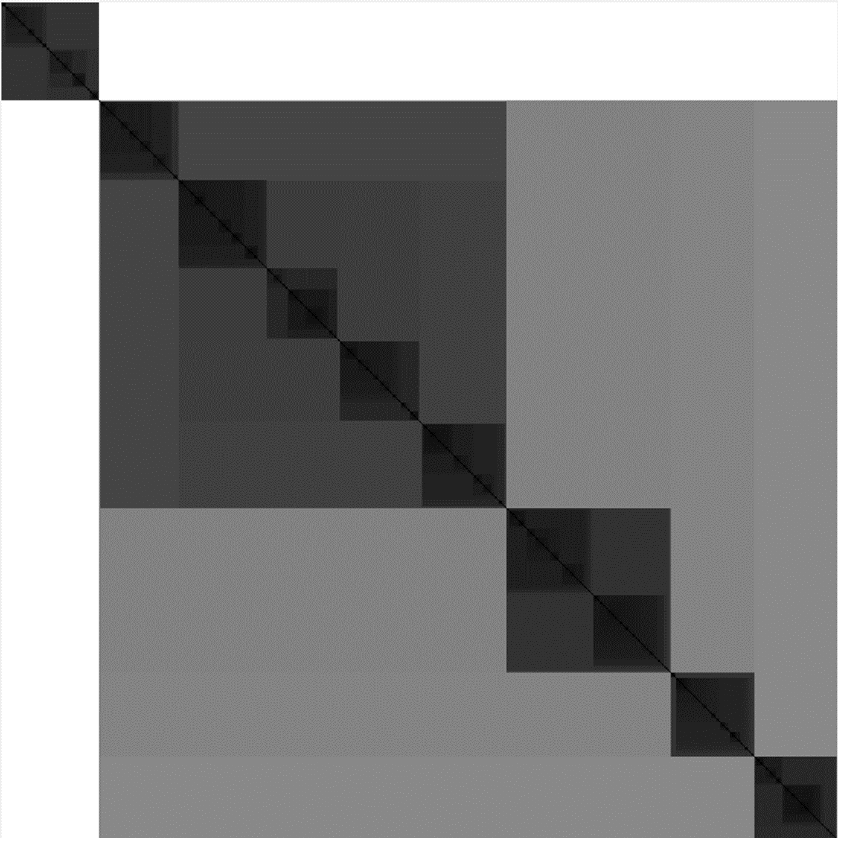}} &
\raisebox{-0.5\height}{\includegraphics[scale=0.2]{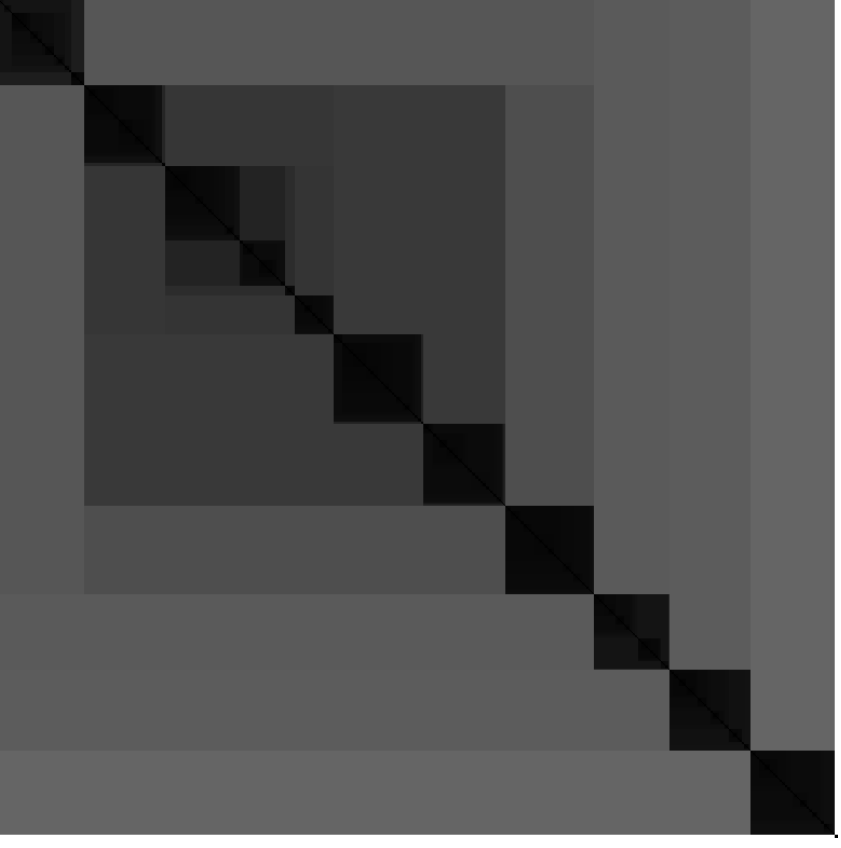}} \\
$k=10$ & $k_{p} = 10$ & $k_{p} = 8$ & $k_{p} = 3$ & $k_{p} = 15$ & $k_{p} = 7$ & $k_{p} = 10$ & $k_{p} = 10$ \\
\hline
\textit{FMNIST} &
\raisebox{-0.5\height}{\includegraphics[scale=0.2]{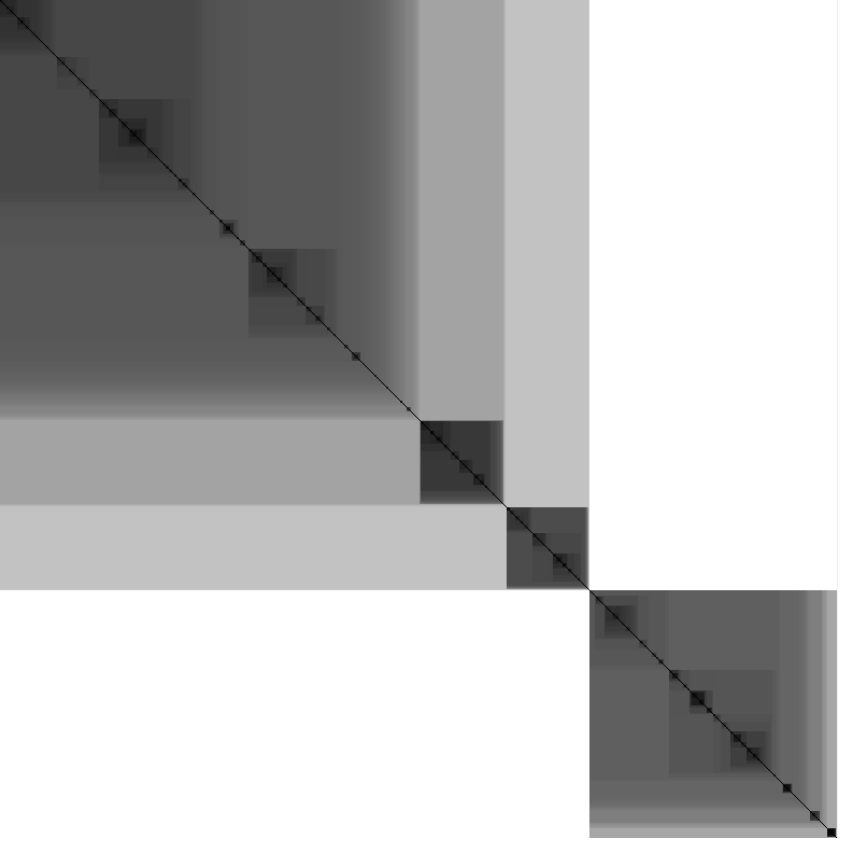}} &
\raisebox{-0.5\height}{\includegraphics[scale=0.2]{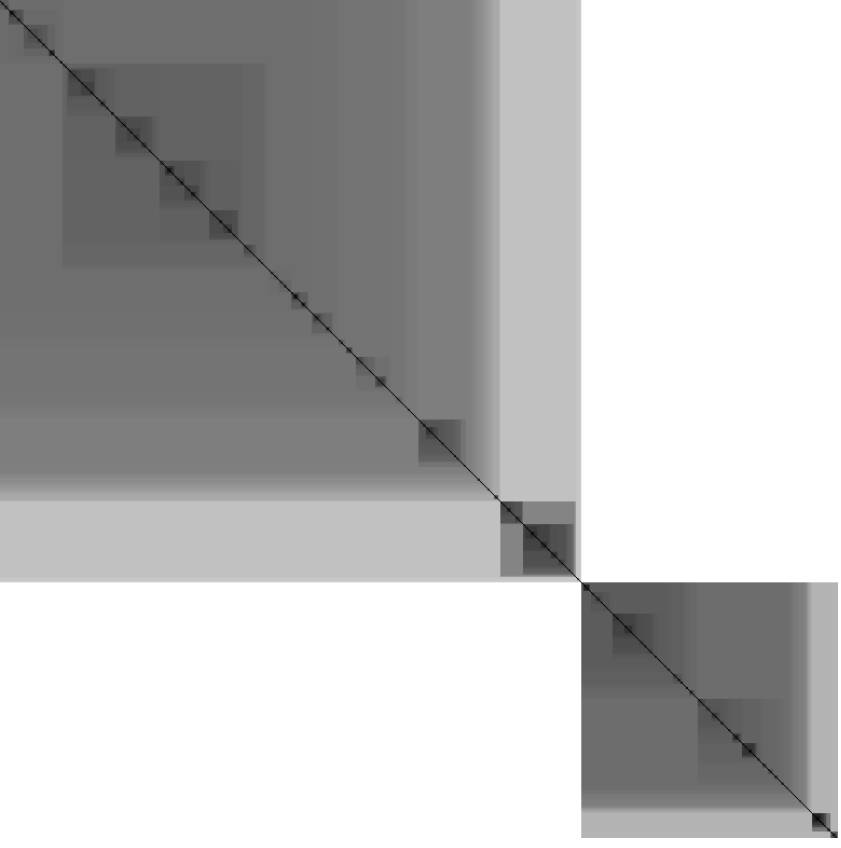}} &
\raisebox{-0.5\height}{\includegraphics[scale=0.2]{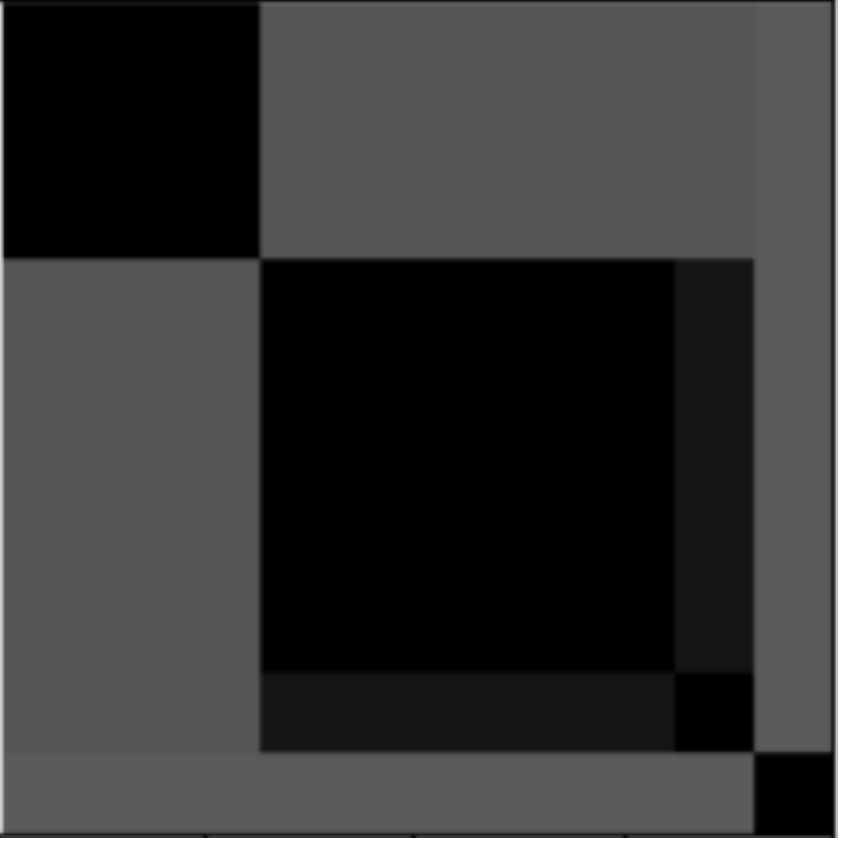}} &
\raisebox{-0.5\height}{\includegraphics[scale=0.2]{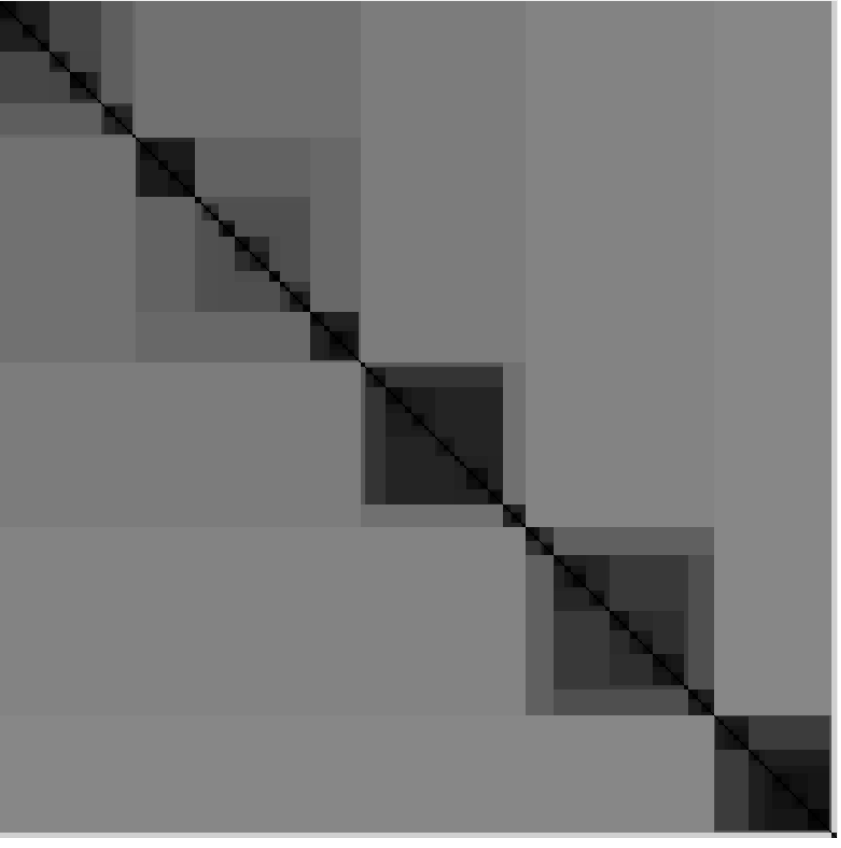}} &
\raisebox{-0.5\height}{\includegraphics[scale=0.2]{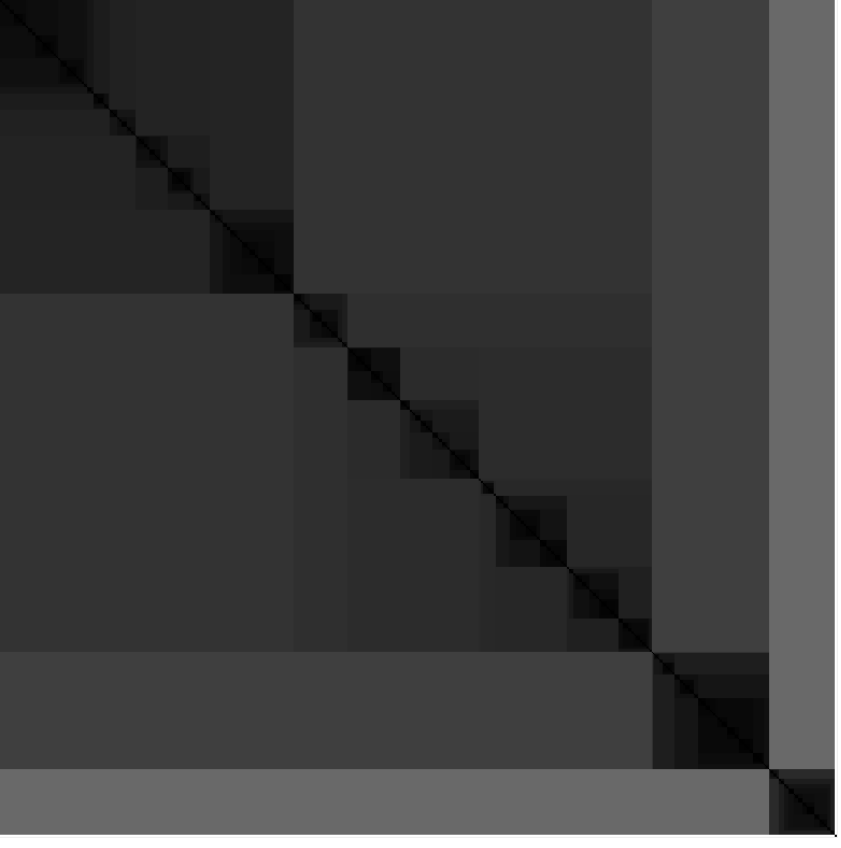}} &
\raisebox{-0.5\height}{\includegraphics[scale=0.2]{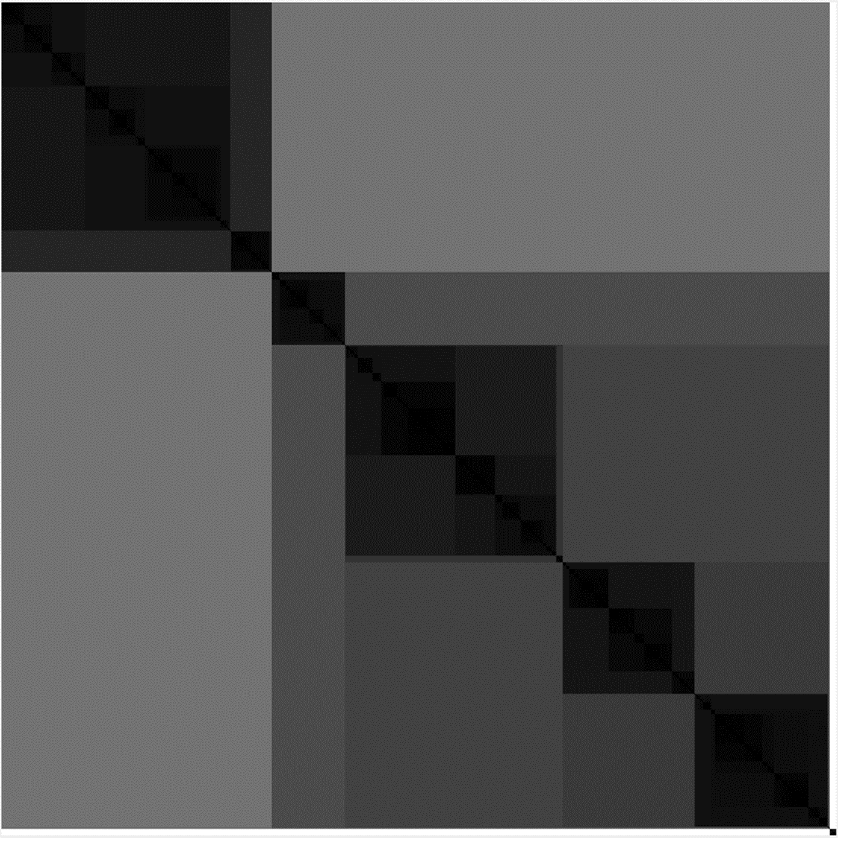}} &
\raisebox{-0.5\height}{\includegraphics[scale=0.2]{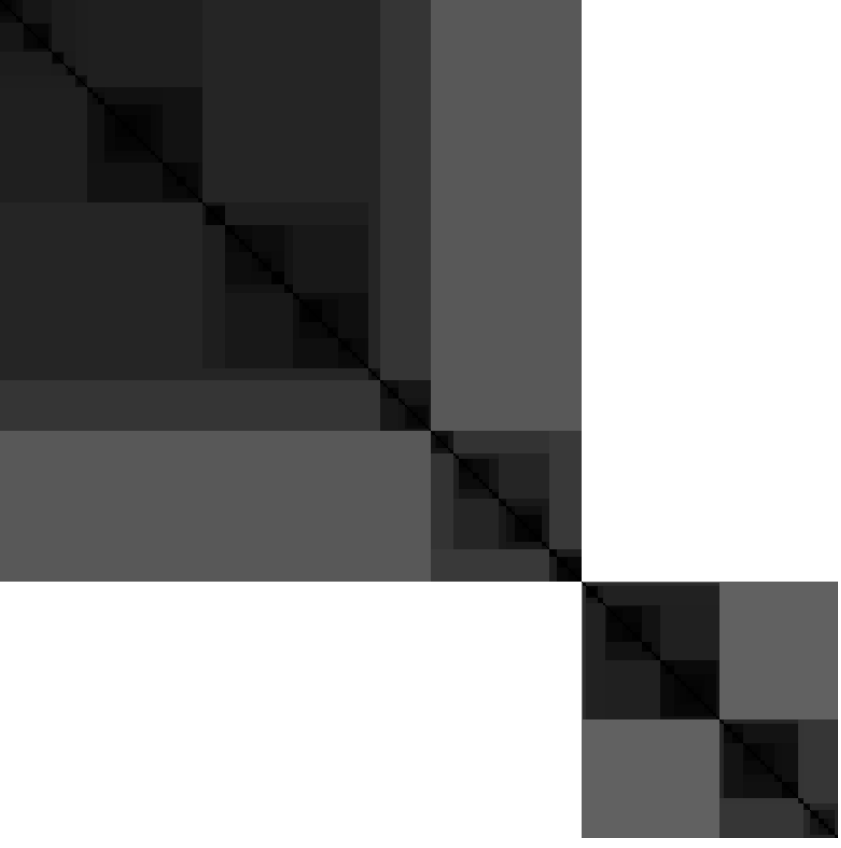}} \\
$k=10$ & $k_{p} = 3$ & $k_{p} = 2$ & $k_{p} = 3$ & $k_{p} = 4$ & $k_{p} = 3$ & $k_{p} = 4$ & $k_{p} = 5$ \\
\hline
\textit{CIFAR-10} &
\raisebox{-0.5\height}{\includegraphics[scale=0.2]{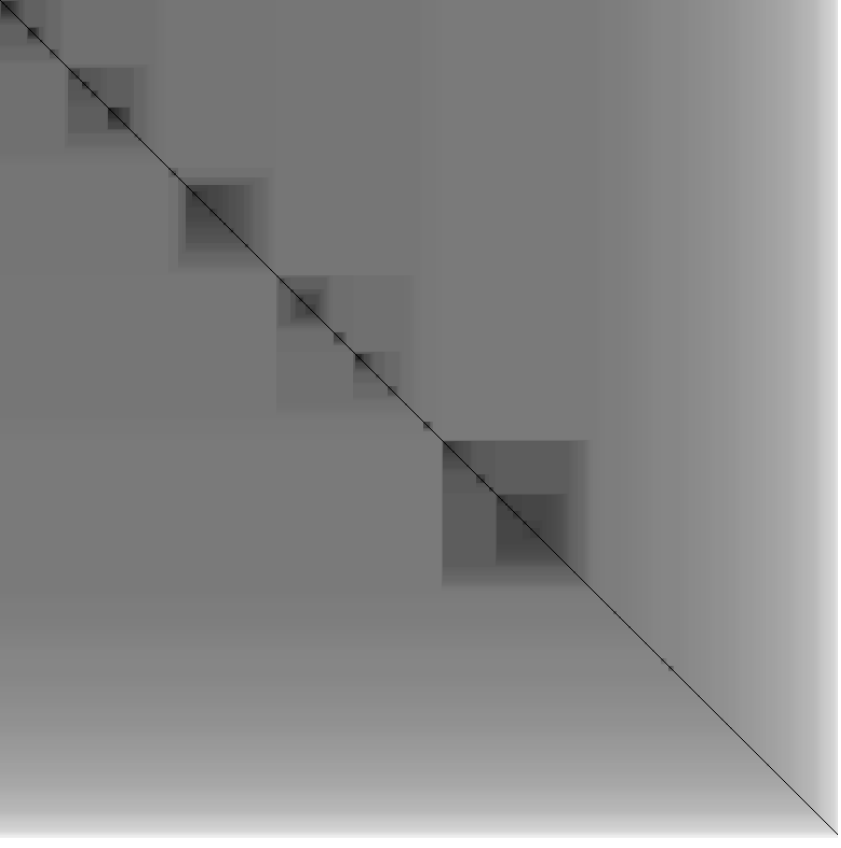}} &
\raisebox{-0.5\height}{\includegraphics[scale=0.2]{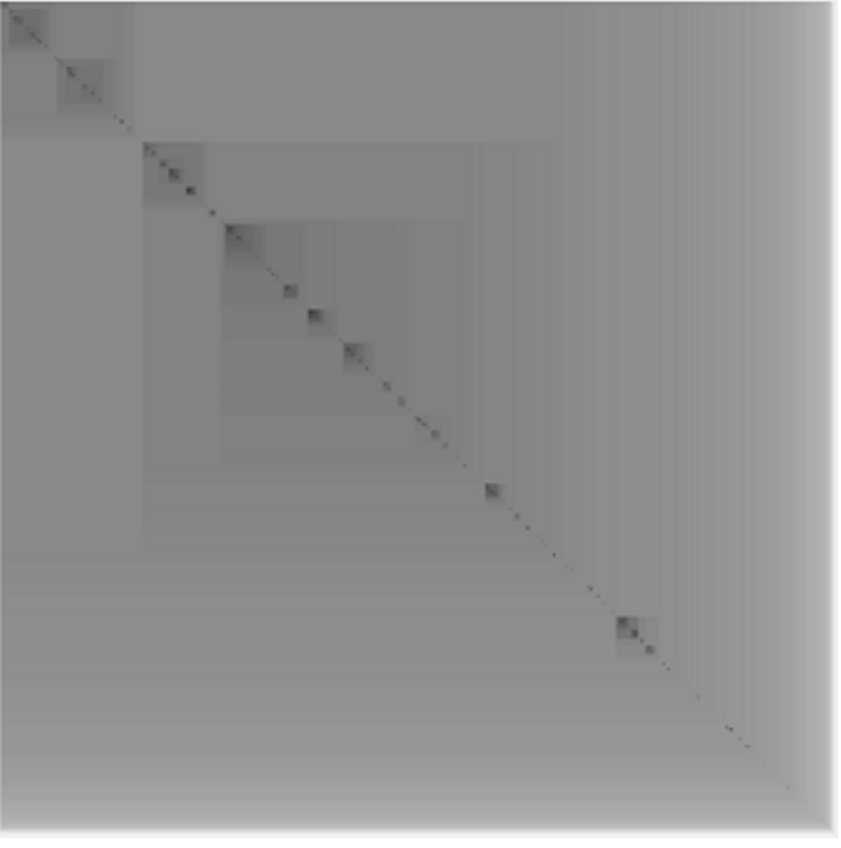}} &
\raisebox{-0.5\height}{\includegraphics[scale=0.2]{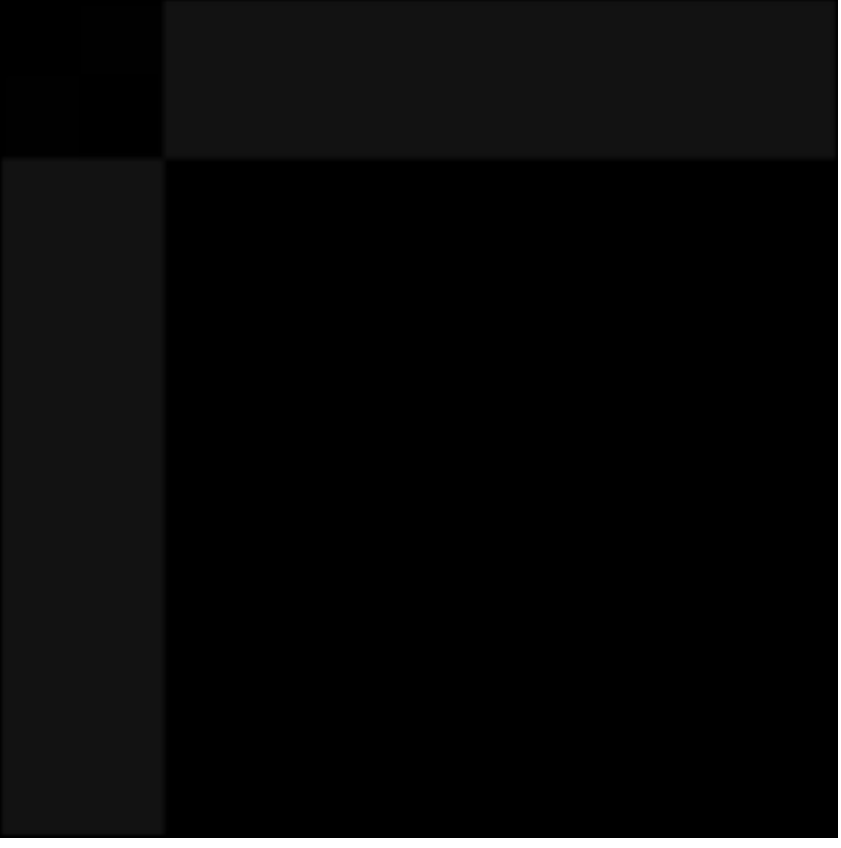}} &
\raisebox{-0.5\height}{\includegraphics[scale=0.2]{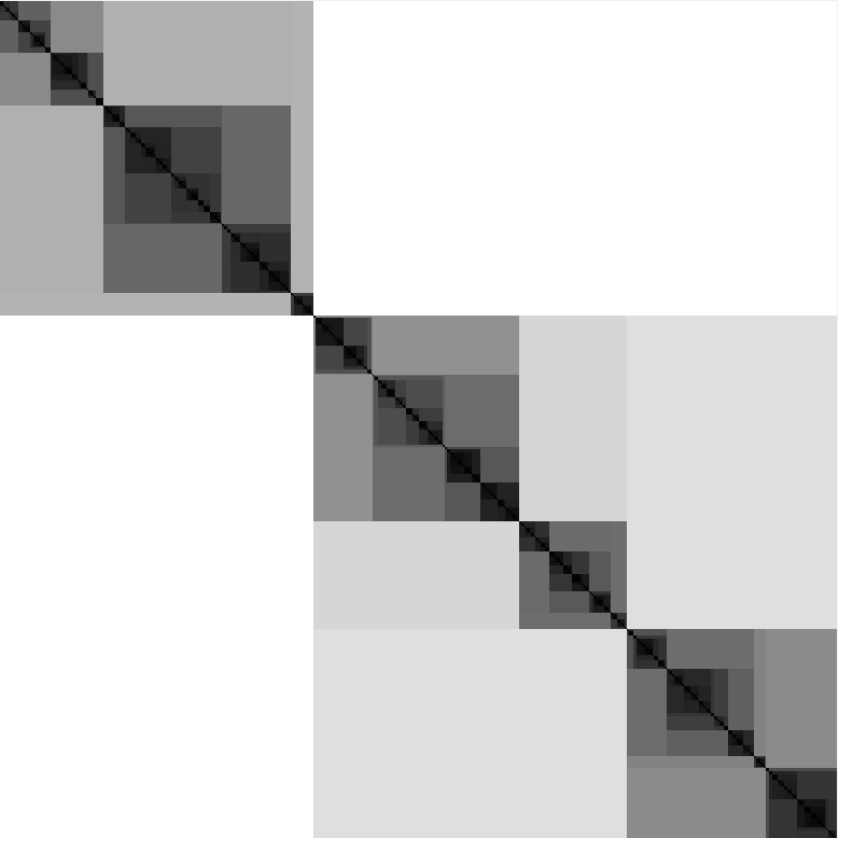}} &
\raisebox{-0.5\height}{\includegraphics[scale=0.2]{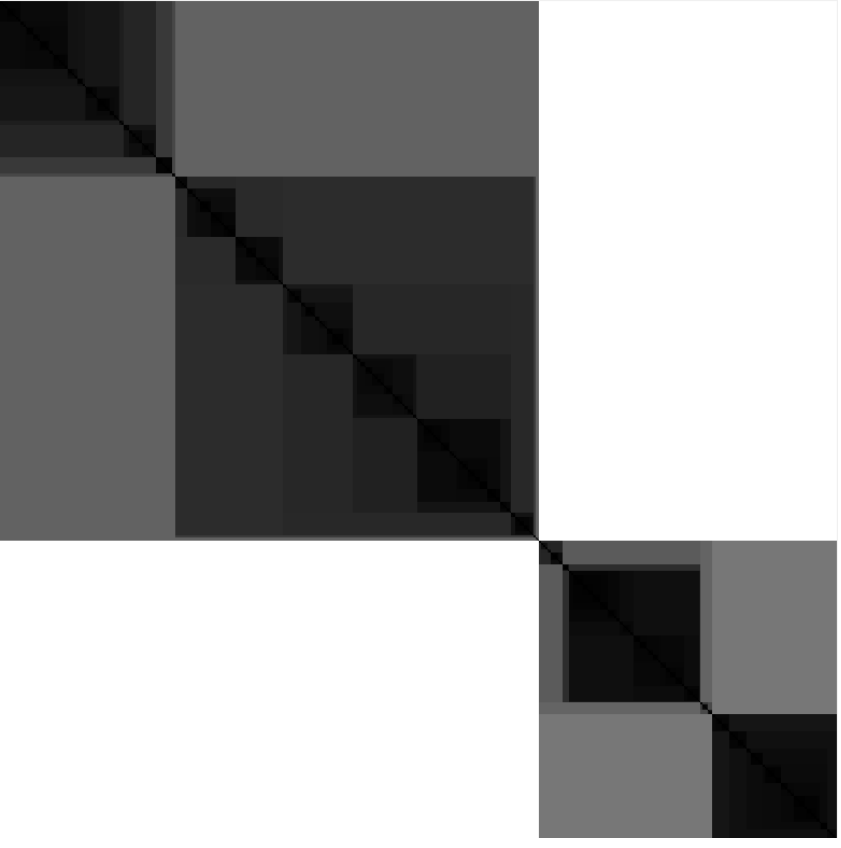}} &
\raisebox{-0.5\height}{\includegraphics[scale=0.2]{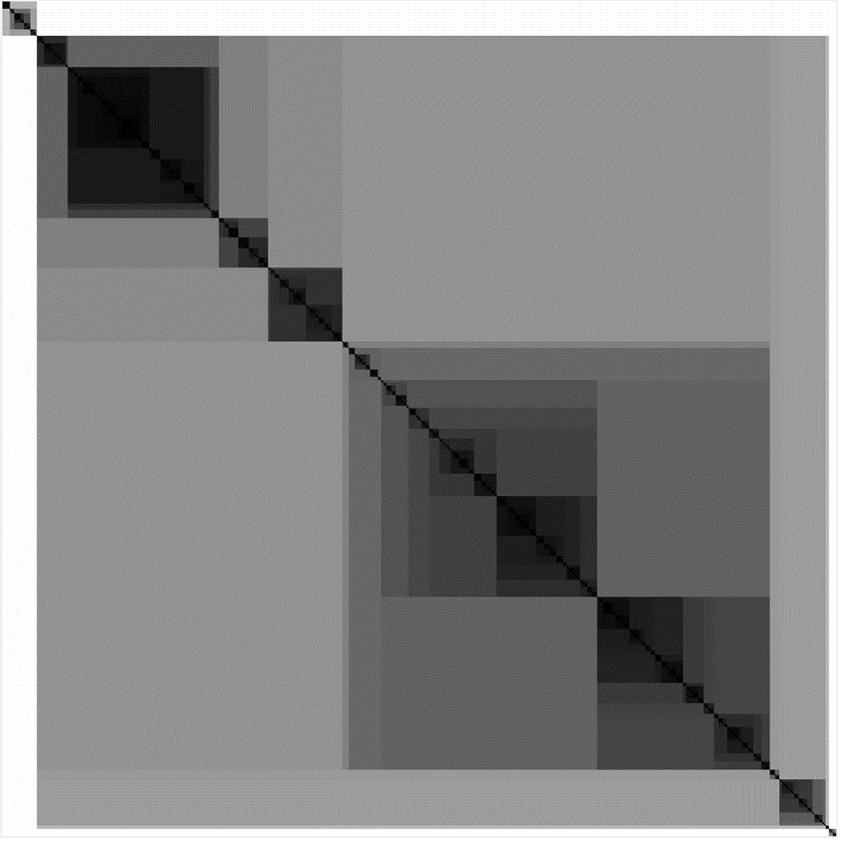}} &
\raisebox{-0.5\height}{\includegraphics[scale=0.2]{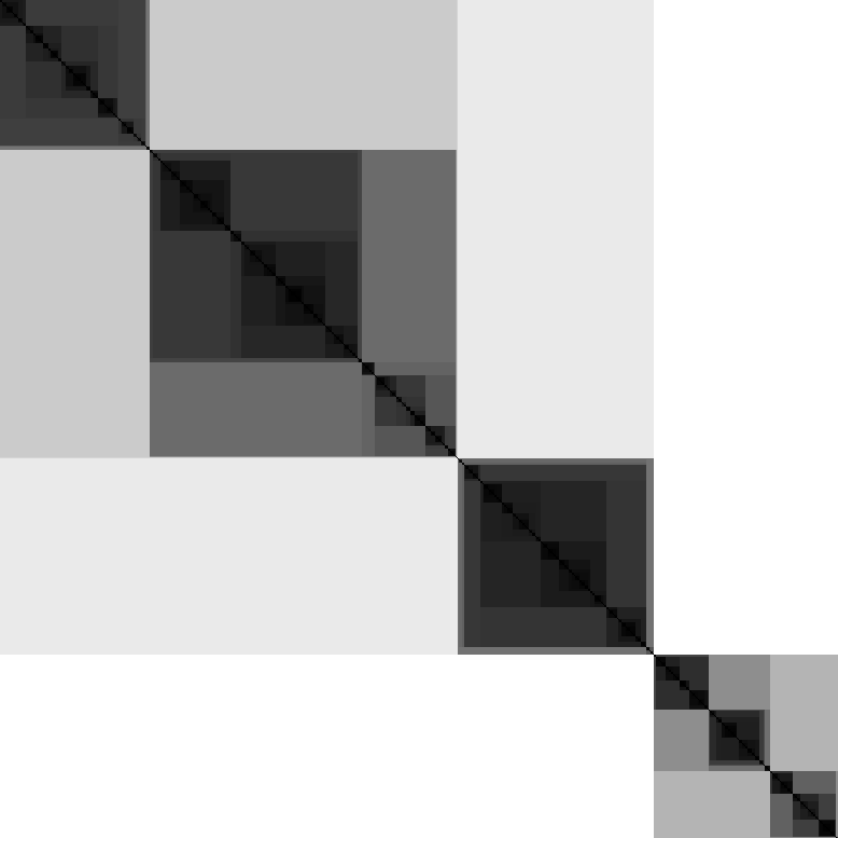}} \\
$k=10$ & $k_{p} = 2$ & $k_{p} = 1$ & $k_{p} = 2$ & $k_{p} = 5$ & $k_{p} = 4$ & $k_{p} = 4$ & $k_{p} = 5$ \\
\hline
\textit{INTEL} &
\raisebox{-0.5\height}{\includegraphics[scale=0.2]{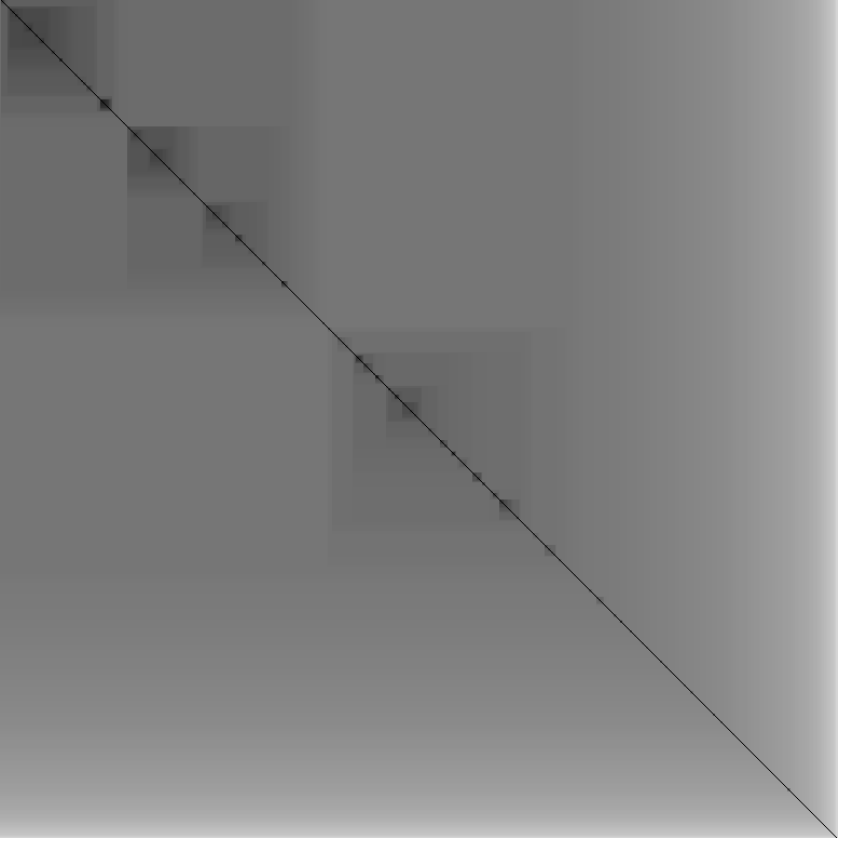}} &
\raisebox{-0.5\height}{\includegraphics[scale=0.2]{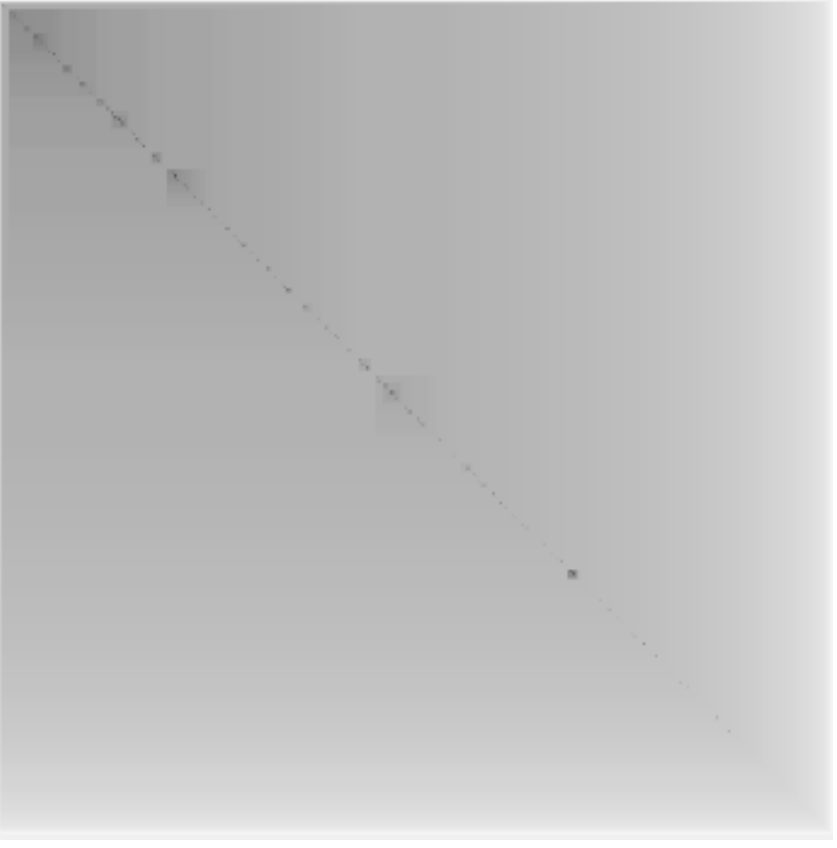}} &
\raisebox{-0.5\height}{\includegraphics[scale=0.2]{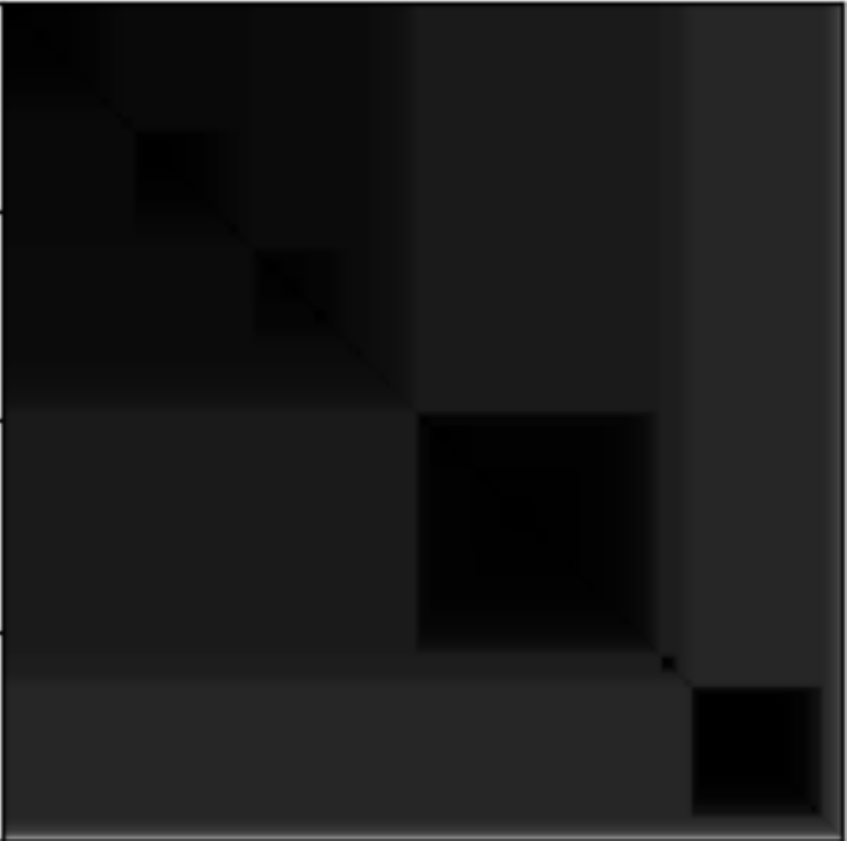}} &
\raisebox{-0.5\height}{\includegraphics[scale=0.2]{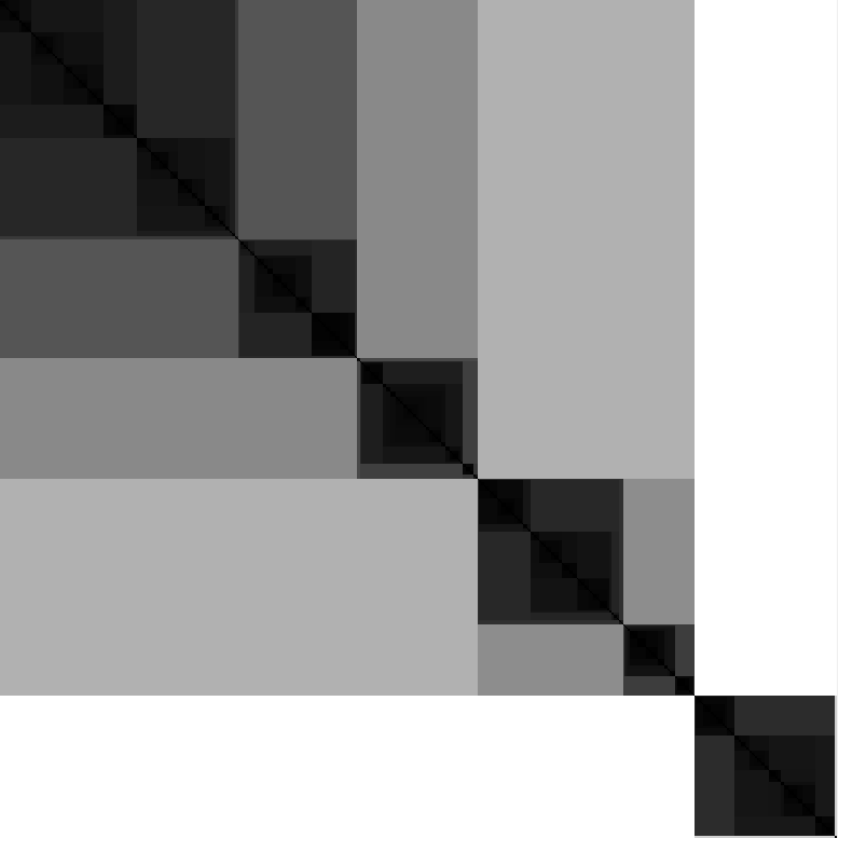}} &
\raisebox{-0.5\height}{\includegraphics[scale=0.2]{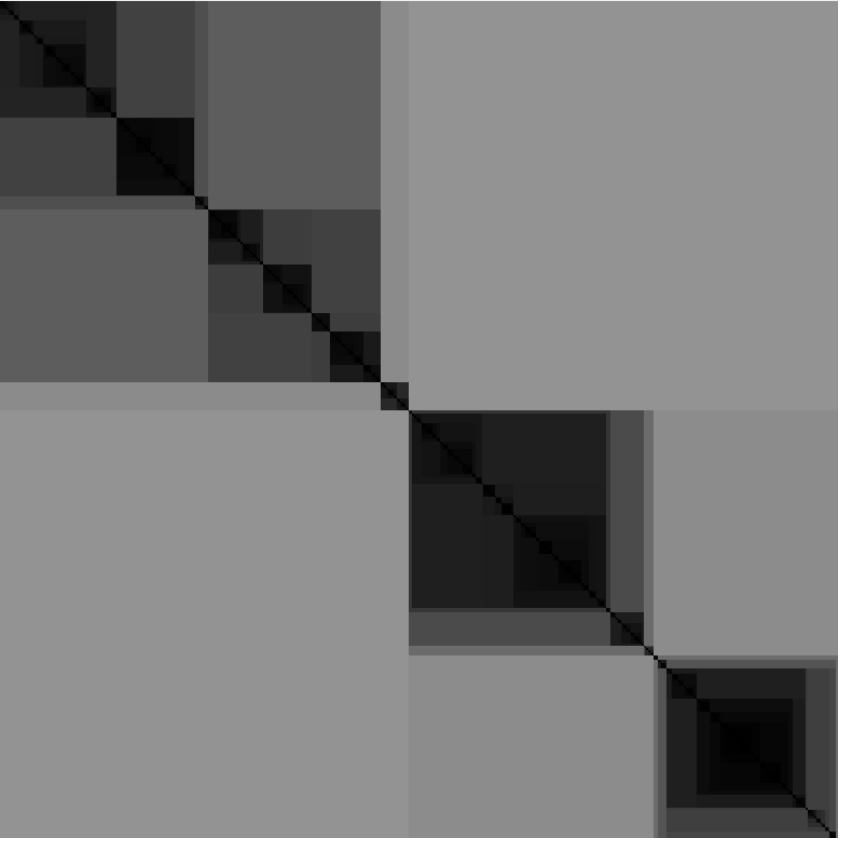}} &
\raisebox{-0.5\height}{\includegraphics[scale=0.2]{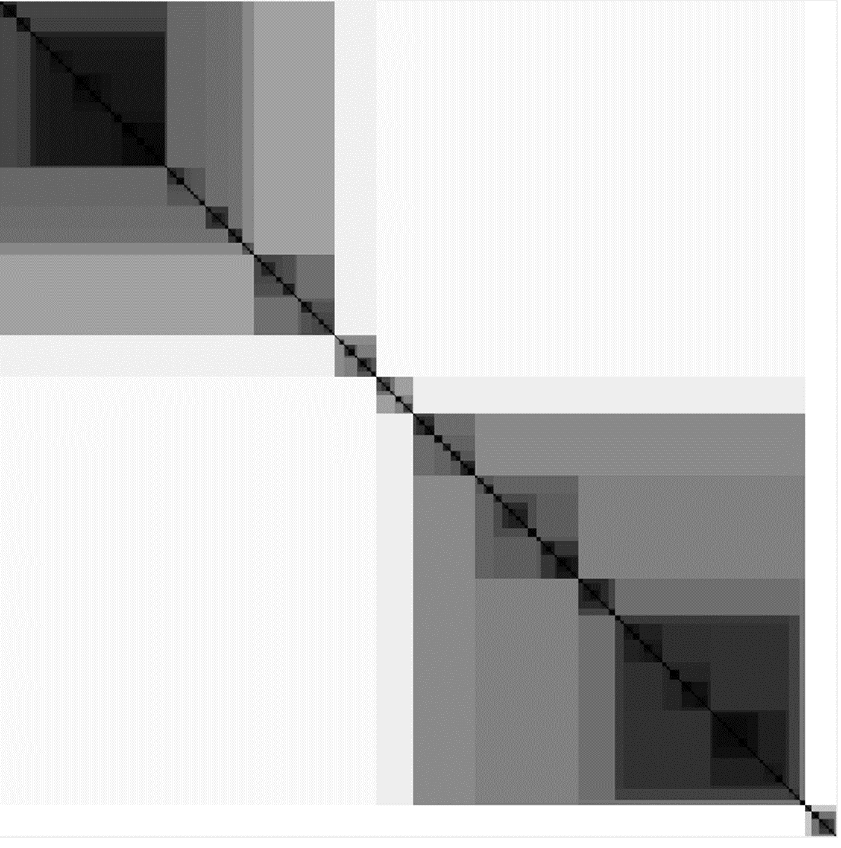}} &
\raisebox{-0.5\height}{\includegraphics[scale=0.2]{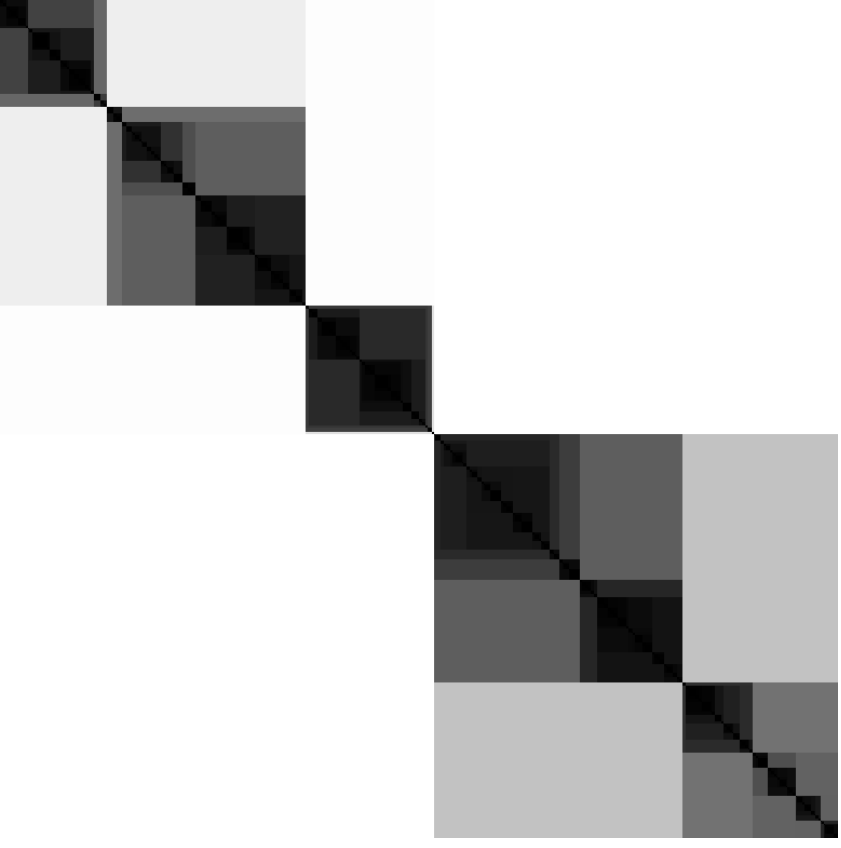}} \\
$k=6$ & $k_{p} = 2$ & $k_{p} = 1$ & $k_{p} = 3$ & $k_{p} = 5$ & $k_{p} = 4$ & $k_{p} = 3$ & $k_{p} = 6$ \\
\hline
\end{tabular}
\end{table*}

\begin{table*}[hbtp!]
\caption{\textbf{Quantitative comparison with other methods.} Our method almost constantly reaches state-of-the-art performances by a good margin on all four datasets.}
\label{tab:resultsB}
\begin{center}
\begin{threeparttable}
\begin{tabular}{|l|cc|cc|cc|cc|}
\hline
\multirow{2}{*}{\backslashbox{\emph{MODELS}}{\emph{DATASETS}}} & \multicolumn{2}{c|}{\emph{MNIST}} & \multicolumn{2}{c|}{\emph{FMNIST}} & \multicolumn{2}{c|}{\emph{CIFAR-10}} & \multicolumn{2}{c|}{\emph{INTEL}} \\
\cline{2-9} 
    & \multicolumn{1}{c|}{PA (\%)} & NMI & \multicolumn{1}{c|}{PA (\%)} & NMI & \multicolumn{1}{c|}{PA (\%)} & NMI & \multicolumn{1}{c|}{PA (\%)} & NMI \\
\hline\hline
\emph{FensiVAT \tnote{\textdagger} on $\Phi$} & \multicolumn{1}{c|}{51.78} & 0.62 & \multicolumn{1}{c|}{29.5} & 0.5 & \multicolumn{1}{c|}{10.2} & 0.01 & \multicolumn{1}{c|}{17.8} & 0.01 \\
\hline
\emph{KernelVAT \tnote{\textdagger} on $\Phi$} & \multicolumn{1}{c|}{40.01} & 0.58 & \multicolumn{1}{c|}{29.4} & 0.49 & \multicolumn{1}{c|}{10.13} & 0.01 & \multicolumn{1}{c|}{17.55} & 0.01 \\
\hline
\emph{SpecVAT  \tnote{\textdagger} on $\Phi$} & \multicolumn{1}{c|}{11.37} & 0.004 & \multicolumn{1}{c|}{10.27} & 0.006 & \multicolumn{1}{c|}{10.19} & 0.004 & \multicolumn{1}{c|}{17.6} & 0.003 \\
\hline\hline
\emph{DEC \tnote{\textdaggerdbl}  + iVAT} & \multicolumn{1}{c|}{51.20} & 0.62 & \multicolumn{1}{c|}{27.69} & 0.22 & \multicolumn{1}{c|}{12.8} & 0.01 & \multicolumn{1}{c|}{31.93} & 0.13 \\
\hline
\emph{LSD-C \tnote{\textdaggerdbl}  + iVAT} & \multicolumn{1}{c|}{51.94} & 0.58 & \multicolumn{1}{c|}{41.47} & 0.46 & \multicolumn{1}{c|}{38.17} & 0.41 & \multicolumn{1}{c|}{39.64} & 0.31 \\
\hline\hline
\emph{Autoencoder + iVAT} & \multicolumn{1}{c|}{41.32} & 0.61 & \multicolumn{1}{c|}{40.80} & 0.53 & \multicolumn{1}{c|}{17.80} & 0.075 & \multicolumn{1}{c|}{30.05} & 0.14 \\
\hline\hline
\textbf{\emph{Ours}} & \multicolumn{1}{c|}{\textbf{82.02}} & \textbf{0.89} & \multicolumn{1}{c|}{\textbf{43.76}} & \textbf{0.61} & \multicolumn{1}{c|}{\textbf{51.26}} & \textbf{0.47} & \multicolumn{1}{c|}{\textbf{56.84}} & \textbf{0.46} \\
\hline
\end{tabular}
\begin{tablenotes}
    \item $\Phi$: 2048 dimensional SimCLR embedding; \textdagger: Models from \emph{VAT/iVAT} family 
    \item \textdaggerdbl: Deep clustering models.
\end{tablenotes}
\end{threeparttable}
\end{center}
\end{table*}

\subsubsection{Parameter Settings}
\label{sec: PS}
In DeepVAT, the SimCLR model was trained using the LARS optimizer~\cite{you2017large} for each dataset, with $1,000$ epochs. The output dimension of the encoder network was set to $d=2,048$, and the projection head network was chosen to have $m=128$. We performed each experiment five times on each dataset and reported the average results. We use a batch size of $700$ for MNIST and FMNIST and $256$ for CIFAR10 and Intel Image Dataset. The parameters for MMRS sampling are $k'$ = 15 for MNIST, FMNIST, and CIFAR10, and  10 for INTEL, number of samples, $n$: 4,000 for all datasets. 

Euclidean distance is utilized as the metric to generate the RDI for the t-SNE reduced embeddings of the MNIST dataset. Likewise, in the ablation study discussed in Section~\ref{sec: ablation}, Euclidean distance is employed to generate the RDI for the t-SNE reduced embeddings of the raw-flattened MNIST data. For all other experiments, cosine dissimilarity is utilized as the dissimilarity measure to generate the RDI.

The input to all three VAT based methods is 2048-dimensional SimCLR embeddings as these methods transform the original data into a suitable embedding space/low dimensional space by virtue of their design. 
In KernelVAT, radial basis function (RBF) kernel is used, with the precision parameter ($\gamma$) set to $0.05$. In FensiVAT, the down-space (reduced) dimension for random projection is chosen $100$ when FensiVAT is applied to a $2048$-dimensional SimCLR embedding. In SpecVAT, we performed iterations over the parameter \textit{number of eigen-values} $(r)$ ranging from 1 to 10 and noted the best result. 


DEC~\cite{pmlr-v48-xieb16} and LSD-C~\cite{Rebuffi_2021_ICCV} heavily rely on prior information about the number of clusters in a dataset, while DeepVAT do not require this specific information. As stated in Section~\ref{sec: PS}, we deliberately choose an overestimate for the number of clusters in all our experiments involving VAT algorithms. Consequently, for a fair comparison, we adopt the same overestimate ($\textit{k}^{'}$ = 15 for MNIST, FMNIST, and CIFAR-10, and $\textit{k}^{'}$ = 10 for INTEL) for DEC (which requires the value of \textit{k} for performing \textit{k}-means) and LSD-C (where the linear layer after the encoder has the same number of neurons as the number of classes).

To ensure fairness in the assessment, just like DeepVAT utilized t-SNE reduced embeddings from the SimCLR encoder, we also apply t-SNE to reduce the embeddings generated from the encoders of DEC and LSD-C to 2 dimensions before generating the RDI.

We keep all the parameter settings the same unless stated otherwise.

\subsection{Ablation Study}
\label{sec: ablation}
\begin{table*}[hbtp!]
\caption{\textbf{Ablation study.} We analyze the effects of removing different blocks from the DeepVAT pipeline on PA and NMI.}
\label{tab:resultsA}
\begin{center}
\begin{tabular}{|l|cc|cc|cc|cc|}
\hline
\multirow{2}{*}{\backslashbox{\emph{MODELS}}{\emph{DATASETS}}} & \multicolumn{2}{c|}{\emph{MNIST}} & \multicolumn{2}{c|}{\emph{FMNIST}} & \multicolumn{2}{c|}{\emph{CIFAR-10}} & \multicolumn{2}{c|}{\emph{INTEL}} \\
\cline{2-9} 
    & \multicolumn{1}{c|}{PA (\%)} & NMI & \multicolumn{1}{c|}{PA (\%)} & NMI & \multicolumn{1}{c|}{PA (\%)} & NMI & \multicolumn{1}{c|}{PA (\%)} & NMI \\
\hline\hline
\textbf{\emph{Full DeepVAT}} & \multicolumn{1}{c|}{\textbf{82.02}} & \textbf{0.89} & \multicolumn{1}{c|}{\textbf{43.76}} & \textbf{0.61} & \multicolumn{1}{c|}{\textbf{51.26}} & \textbf{0.47} & \multicolumn{1}{c|}{\textbf{56.84}} & \textbf{0.46} \\
\hline
\emph{DeepVAT \emph{minus} SimCLR} & \multicolumn{1}{c|}{37.27} & 0.51 & \multicolumn{1}{c|}{29.87} & 0.51 & \multicolumn{1}{c|}{18.25} & 0.07 & \multicolumn{1}{c|}{31.27} & 0.15 \\
\hline
\emph{DeepVAT \emph{minus} t-SNE} & \multicolumn{1}{c|}{40.73} & 0.60 & \multicolumn{1}{c|}{30.61} & 0.52 & \multicolumn{1}{c|}{10.14} & 0.01 & \multicolumn{1}{c|}{17.52} & 0.007 \\
\hline
\emph{DeepVAT \emph{minus} MMRS} & \multicolumn{1}{c|}{--} & -- & \multicolumn{1}{c|}{--} & -- & \multicolumn{1}{c|}{--} & -- & \multicolumn{1}{c|}{--} & -- \\
\hline
\emph{DeepVAT \emph{minus} SimCLR \emph{minus} t-SNE} & \multicolumn{1}{c|}{11.27} & 0.009 & \multicolumn{1}{c|}{10.01} & 0.007 & \multicolumn{1}{c|}{10.12} & 0.06 & \multicolumn{1}{c|}{15.12} & 0.005 \\
\hline
\end{tabular}
\end{center}
\end{table*}

\subsubsection{Results and Discussions}
\label{sec: RD}

Table \ref{tab:resultsC} shows the comparison of all six models based on the RDI quality and their ability to estimate the underlying clusters ($k_p)$ accurately. Table \ref{tab:resultsB} shows the comparison of DeepVAT with all the six methods mentioned above based on the  PA and NMI.

We can see that the DeepVAT method generates much clearer and sharper dark blocks compared to the SpecVAT, KernelVAT, and FensiVAT models. Consequently, the number of dark blocks generated by DeepVAT ($k_p$) is close to the original number of classes ($k$) in the dataset, making it the most accurate in estimating the potential number of clusters compared to other algorithms. When applying FensiVAT and KernelVAT directly to the high-dimensional embedding, we observed that they produced blurry RDI (the cluster count is good but the quality of RDI is poor) and only achieved moderate quantitative results in terms of PA and NMI (refer Table~\ref{tab:resultsB}) for simple datasets such as MNIST and FMNIST. However, when dealing with complex datasets like CIFAR-10 and INTEL, both algorithms failed to generate high-quality RDI and quantitative results (refer Table~\ref{tab:resultsB}). Note that clustering algorithms face significant challenges when dealing with these datasets, as the ground truth labels may not accurately reflect distinct clusters within the feature vector representation of the data points. These results suggest that our approach produces more visually appealing and informative representations of the data.



Based on the results presented in Table~\ref{tab:resultsB}, DeepVAT demonstrates a significant performance advantage over state-of-the-art VAT family methods in terms of both PA and NMI metrics. DeepVAT demonstrates its superiority over deep-clustering algorithms by achieving a 35\% improvement in PA and 30\% improvement in NMI. Furthermore, it outperforms simple autoencoders by an impressive 95\% on PA and 203\% on NMI metrics, clearly highlighting its remarkable performance. As a result, DeepVAT surpasses all six competitive models in both PA and NMI measures.


The success of DeepVAT can be attributed to the use of SimCLR and t-SNE. SimCLR is effective at generating a robust representation of the dataset by leveraging non-linear functions, such as deep CNN encoders and projection heads, to approximate its intrinsic dimensionality. By applying t-SNE on the representation produced by SimCLR, we obtain a better low-dimensional embedding, as SimCLR is better equipped to detect highly varying manifolds than t-SNE alone.

To examine the impact of various components in our model, we perform a three-part ablation study on all four datasets. We systematically eliminate one component at a time from the DeepVAT pipeline (Fig. \ref{fig:method}) to assess its reliance within the complete pipeline. Our model is summarised as \textbf{DeepVAT = SimCLR + t-SNE + MMRS + iVAT}.


\begin{enumerate}
    \item \textbf{DeepVAT \emph{minus} SimCLR}: We flatten each image in the dataset and apply t-SNE on top of them. Then we sample using MMRS and compute the final iVAT image for the samples. This will show that our model not only benefits from the t-SNE block.
    \item \textbf{DeepVAT \emph{minus} t-SNE}: Images are passed through a trained SimCLR encoder, and we sample the learned high-dimensional embeddings using MMRS. The final iVAT image is computed on the sampled embeddings.
    \item \textbf{DeepVAT \emph{minus} MMRS}: iVAT image is computed on full set of embeddings. However, as iVAT/VAT family algorithms require computation of dissimilarity matrix, which has a time complexity of order $\mathcal{O}(N^{2})$, it will take hours to get the results. Hence, due to such large time complexity and resource constraint, we are not reporting the results of this ablation.
    \item \textbf{DeepVAT \emph{minus} tSNE \emph{minus} SimCLR}: We apply iVAT directly on the MMRS sub-set of raw flattened images.
\end{enumerate}


The findings of the ablation study (1) (Table~\ref{tab:resultsA}) suggest that the generation of RDI by DeepVAT is not solely reliant on t-SNE. Although t-SNE applied directly to raw flattened images produces reasonably good results, it is not as accurate as DeepVAT. However, when dealing with complex datasets like CIFAR-10, utilizing t-SNE on raw flattened images fails to provide meaningful information about the cluster structure. Additionally, the role of the SimCLR module in DeepVAT is investigated in the study (2). The results in Table \ref{tab:resultsA} indicate that SimCLR alone does not yield satisfactory outcomes, although it still demonstrates limited interpretability for simple datasets like MNIST and FMNIST. Nevertheless, when iVAT is applied to SimCLR embeddings for complex datasets, it fails to convey meaningful results. This limitation may be attributed to the high dimensionality of the SimCLR embeddings (2048), which hinders the accurate inference of cluster presence by iVAT.

\section{Conclusions and Future Work}
\label{sec:conc}
This article proposes a deep, self-supervised learning based VAT framework, DeepVAT, for cluster structure assessment in image data. The self-supervised learning method SimCLR significantly improved the performance of iVAT both qualitatively and quantitatively. Our experimental results suggest that when t-SNE is used as dimensionality reduction on top of SimCLR embeddings, the iVAT yields a much sharper RDI, thus a more accurate estimate of the number of clusters. This is because SimCLR can capture the intrinsic dimensionality of image datasets which helped t-SNE in generating a good low dimensional representation. Based on our numerical experiments on four image datasets, we have also shown that DeepVAT significantly outperformed other VAT family methods (FensiVAT, KernelVAT and SpecVAT) and two deep clustering methods (DEC and LSD-C) based on clustering partition accuracy (PA) and NMI. We believe that deploying more deep learning based models like deep metric learning and semi-supervised, which have partial access to labels can further improve the iVAT image for complex datasets.

At present, the training time for major self-supervised contrastive learning models is quite extensive. As part of our future work, we aim to focus on reducing the training time required for such models. Our objective is to develop methods that can generate high-quality iVAT images using self-supervised contrastive learning models in significantly less time.

{\small
\bibliographystyle{IEEEtran}
\bibliography{reference.bib}

\begin{thebibliography}{10}
\providecommand{\url}[1]{#1}
\csname url@samestyle\endcsname
\providecommand{\newblock}{\relax}
\providecommand{\bibinfo}[2]{#2}
\providecommand{\BIBentrySTDinterwordspacing}{\spaceskip=0pt\relax}
\providecommand{\BIBentryALTinterwordstretchfactor}{4}
\providecommand{\BIBentryALTinterwordspacing}{\spaceskip=\fontdimen2\font plus
\BIBentryALTinterwordstretchfactor\fontdimen3\font minus
  \fontdimen4\font\relax}
\providecommand{\BIBforeignlanguage}[2]{{%
\expandafter\ifx\csname l@#1\endcsname\relax
\typeout{** WARNING: IEEEtran.bst: No hyphenation pattern has been}%
\typeout{** loaded for the language `#1'. Using the pattern for}%
\typeout{** the default language instead.}%
\else
\language=\csname l@#1\endcsname
\fi
#2}}
\providecommand{\BIBdecl}{\relax}
\BIBdecl

\bibitem{jain1988algorithms}
A.~K. Jain and R.~C. Dubes, \emph{Algorithms for Clustering Data}.\hskip 1em
  plus 0.5em minus 0.4em\relax USA: Prentice-Hall, Inc., 1988.

\bibitem{everitt1978graphical}
B.~S. Everitt, \emph{Graphical techniques for multivariate data}.\hskip 1em
  plus 0.5em minus 0.4em\relax North-Holland, 1978.

\bibitem{cleveland1993visualizing}
W.~S. Cleveland, \emph{Visualizing Data}.\hskip 1em plus 0.5em minus
  0.4em\relax Hobart Press, 1993.

\bibitem{bezdek2002vat}
J.~Bezdek and R.~Hathaway, ``Vat: a tool for visual assessment of (cluster)
  tendency,'' in \emph{Proceedings of the 2002 International Joint Conference
  on Neural Networks. IJCNN'02 (Cat. No.02CH37290)}, vol.~3, 2002, pp.
  2225--2230 vol.3.

\bibitem{prim1957shortest}
R.~C. Prim, ``Shortest connection networks and some generalizations,''
  \emph{The Bell System Technical Journal}, vol.~36, no.~6, pp. 1389--1401,
  1957.

\bibitem{havens2011efficient}
T.~C. Havens and J.~C. Bezdek, ``An efficient formulation of the improved
  visual assessment of cluster tendency (ivat) algorithm,'' \emph{IEEE
  Transactions on Knowledge and Data Engineering}, vol.~24, no.~5, pp.
  813--822, 2012.

\bibitem{rathore2018rapid}
P.~Rathore, D.~Kumar, J.~C. Bezdek, S.~Rajasegarar, and M.~Palaniswami, ``A
  rapid hybrid clustering algorithm for large volumes of high dimensional
  data,'' \emph{IEEE Transactions on Knowledge and Data Engineering}, vol.~31,
  no.~4, pp. 641--654, 2019.

\bibitem{PPR:PPR472596}
\BIBentryALTinterwordspacing
B.~Zhang, Y.~Zhu, S.~Rajasegarar, G.~Li, and G.~Liu, ``Kernel-based ivat with
  adaptive cluster extraction,'' 2022. [Online]. Available:
  \url{https://doi.org/10.21203/rs.3.rs-1483344/v1}
\BIBentrySTDinterwordspacing

\bibitem{vincent2010stacked}
P.~Vincent, H.~Larochelle, I.~Lajoie, Y.~Bengio, P.-A. Manzagol, and L.~Bottou,
  ``Stacked denoising autoencoders: Learning useful representations in a deep
  network with a local denoising criterion.'' \emph{Journal of machine learning
  research}, vol.~11, no.~12, 2010.

\bibitem{chen2020simple}
T.~Chen, S.~Kornblith, M.~Norouzi, and G.~Hinton, ``A simple framework for
  contrastive learning of visual representations,'' in \emph{Proceedings of the
  37th International Conference on Machine Learning}, ser. ICML'20.\hskip 1em
  plus 0.5em minus 0.4em\relax JMLR.org, 2020.

\bibitem{havens2013scalable}
T.~C. Havens, J.~C. Bezdek, and M.~Palaniswami, ``Scalable single linkage
  hierarchical clustering for big data,'' in \emph{2013 IEEE Eighth
  International Conference on Intelligent Sensors, Sensor Networks and
  Information Processing}, 2013, pp. 396--401.

\bibitem{hathaway2006scalable}
\BIBentryALTinterwordspacing
R.~J. Hathaway, J.~C. Bezdek, and J.~M. Huband, ``Scalable visual assessment of
  cluster tendency for large data sets,'' \emph{Pattern Recognition}, vol.~39,
  no.~7, pp. 1315--1324, 2006. [Online]. Available:
  \url{https://www.sciencedirect.com/science/article/pii/S0031320306000550}
\BIBentrySTDinterwordspacing

\bibitem{johnson1990minimax}
\BIBentryALTinterwordspacing
M.~Johnson, L.~Moore, and D.~Ylvisaker, ``Minimax and maximin distance
  designs,'' \emph{Journal of Statistical Planning and Inference}, vol.~26,
  no.~2, pp. 131--148, 1990. [Online]. Available:
  \url{https://www.sciencedirect.com/science/article/pii/037837589090122B}
\BIBentrySTDinterwordspacing

\bibitem{wang2008specvat}
L.~Wang, X.~Geng, J.~Bezdek, C.~Leckie, and R.~Kotagiri, ``Specvat: Enhanced
  visual cluster analysis,'' in \emph{2008 Eighth IEEE International Conference
  on Data Mining}, 2008, pp. 638--647.

\bibitem{zbontar2021barlow}
\BIBentryALTinterwordspacing
J.~Zbontar, L.~Jing, I.~Misra, Y.~LeCun, and S.~Deny, ``Barlow twins:
  Self-supervised learning via redundancy reduction,'' in \emph{Proceedings of
  the 38th International Conference on Machine Learning}, ser. Proceedings of
  Machine Learning Research, M.~Meila and T.~Zhang, Eds., vol. 139.\hskip 1em
  plus 0.5em minus 0.4em\relax PMLR, 18--24 Jul 2021, pp. 12\,310--12\,320.
  [Online]. Available: \url{https://proceedings.mlr.press/v139/zbontar21a.html}
\BIBentrySTDinterwordspacing

\bibitem{avidan2022computer}
\BIBentryALTinterwordspacing
S.~Avidan, G.~Brostow, M.~Ciss{\'{e}}, G.~M. Farinella, and T.~Hassner, Eds.,
  \emph{Computer Vision {\textendash} {ECCV} 2022}.\hskip 1em plus 0.5em minus
  0.4em\relax Springer Nature Switzerland, 2022. [Online]. Available:
  \url{https://doi.org/10.1007/978-3-031-20074-8}
\BIBentrySTDinterwordspacing

\bibitem{chen2021exploring}
X.~Chen and K.~He, ``Exploring simple siamese representation learning,'' in
  \emph{2021 IEEE/CVF Conference on Computer Vision and Pattern Recognition
  (CVPR)}, 2021, pp. 15\,745--15\,753.

\bibitem{grill2020bootstrap}
J.-B. Grill, F.~Strub, F.~Altch\'{e}, C.~Tallec, P.~H. Richemond,
  E.~Buchatskaya, C.~Doersch, B.~A. Pires, Z.~D. Guo, M.~G. Azar, B.~Piot,
  K.~Kavukcuoglu, R.~Munos, and M.~Valko, ``Bootstrap your own latent - a new
  approach to self-supervised learning,'' in \emph{Proceedings of the 34th
  International Conference on Neural Information Processing Systems}, ser.
  NIPS'20.\hskip 1em plus 0.5em minus 0.4em\relax Red Hook, NY, USA: Curran
  Associates Inc., 2020.

\bibitem{oord2018representation}
A.~van~den Oord, Y.~Li, and O.~Vinyals, ``Representation learning with
  contrastive predictive coding,'' 2019.

\bibitem{parulekar2023infonce}
A.~Parulekar, L.~Collins, K.~Shanmugam, A.~Mokhtari, and S.~Shakkottai,
  ``Infonce loss provably learns cluster-preserving representations,''
  \emph{arXiv preprint arXiv:2302.07920}, 2023.

\bibitem{cifar10}
\BIBentryALTinterwordspacing
A.~Krizhevsky, V.~Nair, and G.~Hinton, ``Cifar-10 (canadian institute for
  advanced research),'' 2009. [Online]. Available:
  \url{http://www.cs.toronto.edu/~kriz/cifar.html}
\BIBentrySTDinterwordspacing

\bibitem{intelimages}
\BIBentryALTinterwordspacing
P.~Bansal, ``Intel image classification,'' 2016. [Online]. Available:
  \url{https://www.kaggle.com/datasets/puneet6060/intel-image-classification}
\BIBentrySTDinterwordspacing

\bibitem{van2008visualizing}
\BIBentryALTinterwordspacing
L.~van~der Maaten and G.~Hinton, ``Visualizing data using t-sne,''
  \emph{Journal of Machine Learning Research}, vol.~9, no.~86, pp. 2579--2605,
  2008. [Online]. Available:
  \url{http://jmlr.org/papers/v9/vandermaaten08a.html}
\BIBentrySTDinterwordspacing

\bibitem{lecun-mnisthandwrittendigit-2010}
\BIBentryALTinterwordspacing
Y.~LeCun and C.~Cortes, ``{MNIST} handwritten digit database,''
  http://yann.lecun.com/exdb/mnist/, 2010. [Online]. Available:
  \url{http://yann.lecun.com/exdb/mnist/}
\BIBentrySTDinterwordspacing

\bibitem{xiao2017fashion}
H.~Xiao, K.~Rasul, and R.~Vollgraf, ``Fashion-mnist: a novel image dataset for
  benchmarking machine learning algorithms,'' 2017.

\bibitem{wang2009automatically}
L.~Wang, C.~Leckie, K.~Ramamohanarao, and J.~Bezdek, ``Automatically
  determining the number of clusters in unlabeled data sets,'' \emph{IEEE
  Transactions on knowledge and Data Engineering}, vol.~21, no.~3, pp.
  335--350, 2009.

\bibitem{lovasz1986matching}
L.~Lov{\'a}sz and M.~Plummer, ``Matching theory, akad,'' \emph{Kiad{\'o},
  Budapest}, 1986.

\bibitem{pmlr-v48-xieb16}
J.~Xie, R.~Girshick, and A.~Farhadi, ``Unsupervised deep embedding for
  clustering analysis,'' in \emph{Proceedings of the 33rd International
  Conference on International Conference on Machine Learning - Volume 48}, ser.
  ICML'16.\hskip 1em plus 0.5em minus 0.4em\relax JMLR.org, 2016, p. 478–487.

\bibitem{Rebuffi_2021_ICCV}
\BIBentryALTinterwordspacing
S.~Rebuffi, S.~Ehrhardt, K.~Han, A.~Vedaldi, and A.~Zisserman, ``Lsd-c:
  Linearly separable deep clusters,'' in \emph{2021 IEEE/CVF International
  Conference on Computer Vision Workshops (ICCVW)}.\hskip 1em plus 0.5em minus
  0.4em\relax Los Alamitos, CA, USA: IEEE Computer Society, oct 2021, pp.
  1038--1046. [Online]. Available:
  \url{https://doi.ieeecomputersociety.org/10.1109/ICCVW54120.2021.00121}
\BIBentrySTDinterwordspacing

\bibitem{you2017large}
Y.~You, I.~Gitman, and B.~Ginsburg, ``Large batch training of convolutional
  networks,'' 2017.

\end{thebibliography}
}

\end{document}